\documentclass[11pt]{article}

\usepackage[final]{acl}

\usepackage{times}
\usepackage{latexsym}
\usepackage[T1]{fontenc}
\usepackage[utf8]{inputenc}
\usepackage{microtype}
\IfFileExists{inconsolata.sty}{\usepackage{inconsolata}}{\usepackage{courier}}
\usepackage{amsmath,amssymb}
\usepackage{fancyvrb}
\usepackage{graphicx}
\usepackage{booktabs}
\usepackage{float}
\usepackage{placeins}
\usepackage{naacl_float_placement}
\usepackage{fvextra}
\usepackage{xcolor}
\usepackage[most]{tcolorbox}
\usepackage{listings}

\title{Narrative Lookahead in LLMs: Event-Level Decodability, Predictive Advantage, and Steering}

\author{
  Donghoon Jung\thanks{Equal contribution.}\quad 
  Min Kim\footnotemark[1]\quad 
  Seojin Choi\footnotemark[1]\quad 
  Songeun Chae \quad 
  Seohyon Jung\thanks{Corresponding author.} \\
  School of Digital Humanities and Computational Social Sciences, KAIST, South Korea \\
  \texttt{\{donghoon.jung, mk, seojin64, songeun, seohyon.jung\}@kaist.ac.kr} 
}

\begin{document}
\maketitle
\begin{abstract}
Do language models carry information about narrative events before writing them, and how far ahead? We examine narrative lookahead through salient, state-changing events rather than tokens or response-level attributes. We extract 23,355 LLM-annotated events from 2,000 stories generated by four open-weight LLMs, labeling them along three nonexclusive dimensions of change: within agents, between agents, and in the story world, with human validation on a subset. Event segmentation and plot-independent labels enable comparisons across stories at horizons measured in events rather than in tokens. At event ends, we probe hidden states for upcoming event types and compare predictions with text-only predictors using matched prefixes. Across four models, future event types are decodable above prevalence baselines for at least four subsequent events, but probes significantly outperform both primary text-only predictors only for the next event. In exploratory steering with OLMo-3.1-32B, a probe-derived direction shifts the proportion of the target event type in both intended directions at higher strengths. Our event-level approach links the prediction of future narrative developments to interventions that reshape subsequent event composition.
\end{abstract}

\section{Introduction}
How far ahead language models look during generation depends in part on what we count as the future. Studies of hidden states have recovered information about subsequent tokens and properties of eventual responses \citep{pal2023future,dong2025emergent}. Yet the future of a story is not exhausted by its forthcoming wording or final form. It takes shape through events that change characters, relationships, and the story world. Storytelling thus offers a way to ask how much information about successive narrative developments is decodable from hidden states before they are written, a question that bears on how we interpret and control long-form generation.

\begin{figure}[!t]
    \centering
    \includegraphics[width=\columnwidth]
    {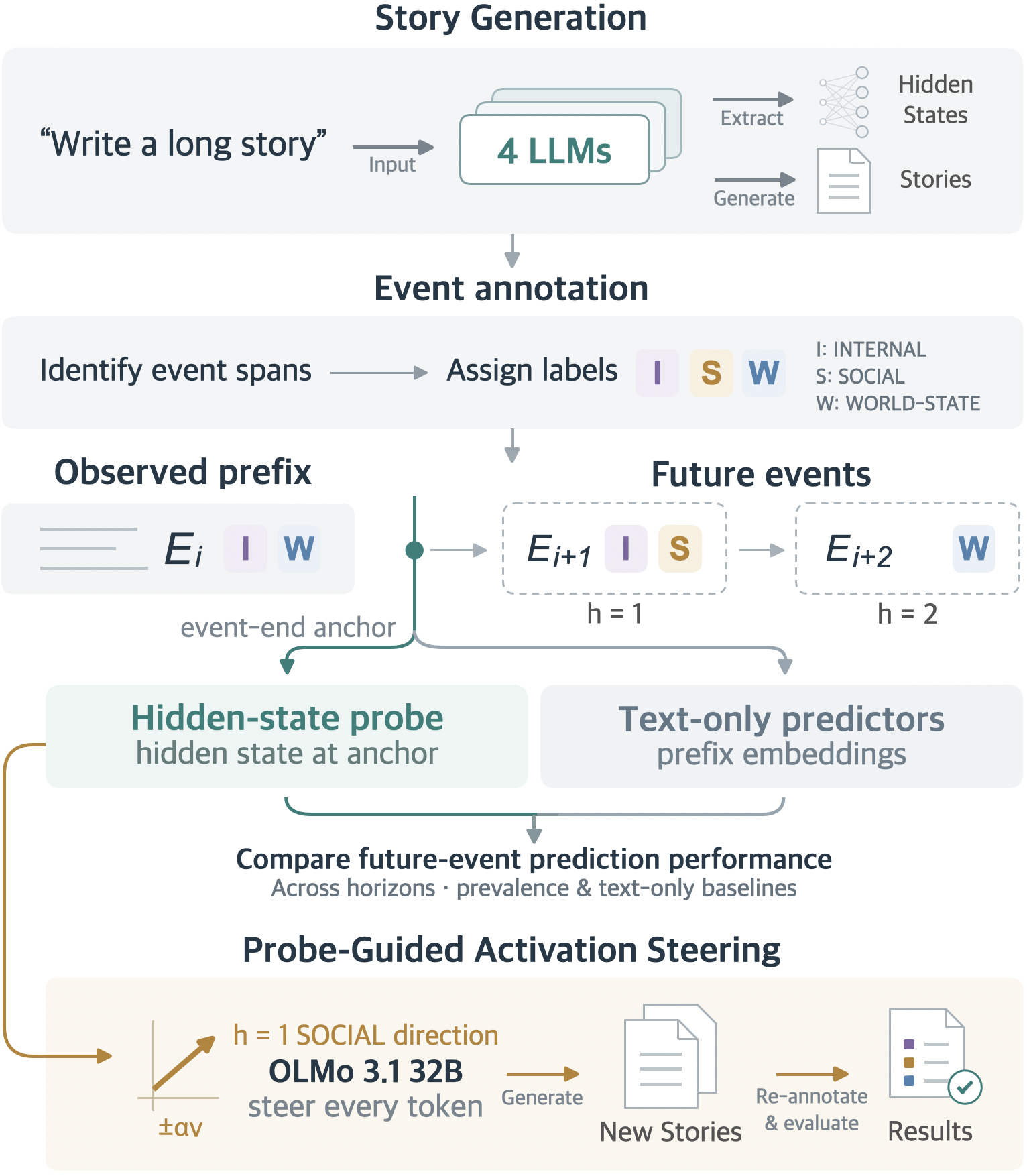}
    \caption{Overview of the research workflow. Generated stories
    are annotated with event spans and nonexclusive state-change
    labels. Hidden-state and matched text-only predictors target
    not-yet-started events across horizons. A next-event SOCIAL
    probe guides sustained steering of OLMo-3.1-32B-Instruct from the
    original prompt. New stories are re-annotated to evaluate
    event composition and generation-quality diagnostics.}
    \label{fig:overview}
\end{figure}

To compare future developments across stories with different plots, we characterize narrative events not by plot-specific content but by where change occurs: within an agent (INTERNAL), between agents (SOCIAL), or in the story world (WORLD-STATE). These plot-independent categories make upcoming event types comparable across stories, while representing stories as ordered event sequences allows lookahead to be measured in successive events rather than tokens. Yet decodability alone cannot tell us whether hidden-state probes offer a predictive advantage over representations of the preceding text or whether directions derived from future-event prediction can influence subsequent generation. We therefore examine the horizon of event-type decodability, compare hidden-state probes with text-only predictors using matched prefixes, and test whether probe-guided steering changes subsequent event composition.

We generate 2,000 stories (500 per model) using four 24–32B open-weight, instruction-tuned LLMs, each given the same minimal user instruction, Write a long story. An LLM annotator extracts 23,355 salient events and assigns the three nonexclusive labels (INTERNAL, SOCIAL, WORLD-STATE). We validate event selection and state-change labels against judgments from three human annotators on 30 OLMo-generated stories. At event ends, we probe hidden states and compare their predictions with those of predictors trained on text-only representations of the same prefix. We then conduct exploratory sustained activation steering in OLMo-3.1-32B-Instruct.

Our contributions are threefold. First, we introduce an event-level framework for studying prospective information in open-ended generation, in which plot-independent state-change labels make lookahead measurable in narrative events. Second, we separate decodability from predictive advantage. Across all four models, future event types are decodable above prevalence baselines for at least four subsequent events, with gains generally declining with distance, yet hidden-state probes significantly outperform both full-prefix and chunk-averaged text-embedding predictors only for the next event. Third, we use event composition as an intervention outcome. In OLMo, sustained steering along a next-event SOCIAL probe direction shifts the proportion of SOCIAL events in both intended directions at higher strengths, with some generation-quality costs, whereas Gemma, despite comparably strong next-event decodability, shows no comparable control under the tested protocol.

\section{Related Work}
\subsection{Future Information in LLMs}
Prior work has shown that LLM hidden states encode information about outputs not yet generated. \citet{pal2023future} recover information about multiple subsequent tokens from individual hidden states, while \citet{men-etal-2024-unlocking} identify representations of short-term future decisions in planning tasks. \citet{zhang2025reasoningmodelsknowtheyre} show that hidden states can predict the correctness of future answers during reasoning, before those answers are fully formulated. More closely related to narrative generation, \citet{dong2025emergent} demonstrate that prompt representations predict response-level attributes, including response length, reasoning steps, and character choices in generated stories. They also examine how these signals evolve during generation.

A related line of work examines how far into the future such information remains decodable. \citet{xu2026faraheadllmsplan} investigate future-information horizons during chain-of-thought reasoning, while \citet{ustaomeroglu2026internalplanninglanguagemodels} characterize the planning horizon of pre-output computations across different tasks. Their findings suggest that the extent of decodable future information depends on the task and prediction distance. These studies leave open how far future-event information can be decoded across successive narrative developments during open-ended story generation.

\subsection{Event-Based Narrative Modeling and Generation}
Narrative events provide a level of representation beyond individual textual units. Narratology characterizes events as changes of state and distinguishes consequential developments from less significant occurrences \citep{huhn2013event}. In NLP, narrative event chains have been used to model relationships and likely continuations among events \citep{chambers-jurafsky-2008-unsupervised,Granroth2016happens}, while research on literary event detection and salience has examined how to identify narratively significant developments \citep{sims-etal-2019-literary,otake-etal-2020-modeling,wilmot-keller-2021-memory,zhang-etal-2021-salience,sharma-etal-2026-upon}.

Event representations have also informed automatic story generation. \citet{Martin_2018} use event sequences as intermediate semantic representations before realizing them as text, while other systems guide generation explicitly through storylines, outlines, or event-based plot structures \citep{yao2019plan,rashkin-etal-2020-plotmachines,yang2023doc,li2025storyteller}. Across these approaches, events serve as structures to be identified, modeled, or used to guide generation. Our study instead uses plot-independent event types as targets of prospective prediction and as outcome measures for intervention, without an externally supplied plot.

\subsection{Probing and Representation Intervention}
Probing can reveal information in hidden states without establishing its role in generation. Prior work has examined the interpretive limitations of probes \citep{hewitt2019designinginterpretingprobescontrol,belinkov-2022-probing} and used representational interventions to test behavioral relevance \citep{elazar-etal-2021-amnesic}. Causal approaches likewise emphasize connecting internal representations to model behavior \citep{mueller-etal-2026-quest}. These distinctions matter when interpreting future information as evidence of planning: decodable information may reflect anticipation or information useful for current predictions \citep{wu2024languagemodelsplanahead}. Stronger causal accounts require effects on both a future target and the context preparing for it \citep{hanna2026latent}, while future-rhyme interventions reveal effects that vary by model and intervention position \citep{ma2026where}.

Representation engineering and activation steering offer methods for manipulating hidden states and examining changes in generation \citep{zou2023representation,turner2023activation,rimsky-etal-2024-steering}. Yet intervention effects raise questions of effectiveness, faithfulness, and selectivity \citep{stoehr-etal-2024-activation,canby-etal-2025-reliable}. We therefore distinguish future-event decodability from steering-induced changes in event composition to examine both the predictive accessibility of future-event information and its potential role in narrative generation.

\section{Methodology}
We construct the dataset by generating stories and representing them as annotated event sequences. We then conduct three complementary analyses: activation probing, comparison with text-only predictors, and probe-guided activation steering.

\subsection{Story Generation and Hidden-State Extraction}
\label{sec:story-generation}

\paragraph{Models and sampling.}
We generate 500 stories with each of four open-weight, dense, instruction-tuned models: OLMo-3.1-32B-Instruct \citep{olmo2026olmo3}, Qwen3.5-27B \citep{qwen35modelcard}, Gemma-4-31B-it \citep{gemmateam2026gemma4technicalreport}, and Mistral-Small-3.2-24B-Instruct-2506 \citep{mistralsmall32}. We refer to these models as OLMo, Qwen, Gemma, and Mistral, respectively.

This 24B--32B range limits size variation across model families. We focus on larger models given reported scale effects in the decodability of future-output information \citep{dong2025emergent,ma2026where} and in latent planning \citep{hanna2026latent}. Each model receives a fixed user prompt \texttt{Write a long story} through its native chat template, without an explicitly supplied system message; OLMo's template-provided default is retained. Thinking modes are disabled for Qwen and Gemma. Sampling uses temperature 0.9, top-$p=0.95$, no top-$k$ filtering, and no repetition penalty. We accept only outputs ending with a configured end-of-sequence (EOS) token within 500--4,000 generated tokens, counting EOS but excluding the prompt. These criteria exclude short or truncated outputs to retain narratives of sufficient length for our analysis. Rejected attempts are replaced using new seeds until 500 stories are accepted per model. Checkpoints and generation details appear in Appendix~\ref{app:generation-details}.

\paragraph{Hidden-state extraction.}
During autoregressive generation with key--value (KV) caching, we save hidden states from every transformer layer, excluding the embedding output, at the final prompt token and candidate sentence boundaries. Each boundary state is measured after processing the token completing its final character. Later tokens may confirm a boundary but do not change the saved state. Event ends are subsequently aligned with these candidates; anchors that contain subsequent narrative text are excluded. Boundary rules and alignment criteria are given in Appendices~\ref{app:boundary-details} and~\ref{app:prediction-details}.

\subsection{Narrative Event Representation}
\label{sec:event-representation}

\paragraph{Main events.}
We represent each story as an ordered sequence of consequential narrative developments, following the distinction between salient events and supporting material in narrative research \citep{otake-etal-2020-modeling,wilmot-keller-2021-memory,zhang-etal-2021-salience}. An event introduces a meaningful new narrative state; static descriptions, minor actions, and intermediate steps are not extracted independently. Closely connected changes constituting one development are grouped, while distinct developments remain separate. Spans comprise the smallest set of consecutive sentence units needed to capture an event. Event counts vary across stories, and future horizons index successive events rather than fixed textual distances.

\paragraph{State-change attributes.}
We characterize events by where narrative change occurs rather than by particular actions or plot functions. Informed by narratological work on narrative experience, socially embedded fictional minds, and storyworlds \citep{fludernik1996natural,palmer2010social,ryan1991possible}, we define three nonexclusive attributes: INTERNAL (I), a meaningful cognitive, affective, or motivational change within a character; SOCIAL (S), a change in interpersonal relations or social structure; and WORLD-STATE (W), a change in the external storyworld, including physical, environmental, or possession states. At least one attribute must be present, and multiple attributes may apply to one event. The rubric also included salient attempts at a state change, with attributes assigned according to the intended domain even when the change was not realized. Because such Attempted events were rare (0.9–2.0\% of events across models), Attempted/Realized status was retained as metadata rather than used as a prediction target.

\paragraph{Annotation.}

 We annotate each complete story independently with \texttt{gpt-6-sol}, using a fixed two-stage rubric that first identifies main events and then assigns state-change labels. Annotations are returned as structured JSON, with event spans required to be verbatim and contiguous. Partial-sentence spans are deterministically expanded to sentence boundaries, retaining their original labels and recording the original spans. Stories with event numbering inconsistent with narrative order were regenerated and reannotated. API settings, validation, and retry policies appear in Appendix~\ref{app:annotation-details}; the full rubric is reproduced in Appendix~\ref{app:annotation-prompt}. Three human annotators independently assessed the extracted events and state-change labels using 30 OLMo stories from the main dataset (Appendix~\ref{app:human-validation}).

\subsection{Activation Probing}
\label{sec:activation-probing}

We test whether frozen hidden states predict narrative event types before the events begin. At an event-end anchor, horizon $h$ denotes the $h$th event that has not yet started, whose INTERNAL, SOCIAL, and WORLD-STATE attributes are predicted independently. Anchors with fewer than $h$ remaining future events are excluded at that horizon, since the target event does not exist and should not be treated as a negative instance.

For each model, we divide its 500 stories into 300 training, 100 validation, and 100 test stories, keeping all samples from a story together. We fit an independent $\ell_2$-regularized logistic probe for each layer, horizon, and attribute \citep{gao2026raptor}, selecting regularization by story-balanced validation average precision (AP). We fix one representative layer per model by maximizing mean event-end validation Macro-AP across eligible horizons, where Macro-AP averages the three attribute APs.

Test predictions are pooled within each horizon with equal total weight per story and compared against constant training-prevalence scores. Paired story bootstrap intervals and within-model Holm correction assess AP differences. Anchor construction, fitting details, and time-resolved analyses are specified in Appendix~\ref{app:prediction-details}.

\subsection{Comparison with Text-Only Predictors}
\label{sec:observable-text}

We compare activation probes with text-only predictors using the same story splits, targets, horizons, and probe-fitting procedure. Text inputs end at the activation's exact character cutoff and exclude the generation prompt. We use \texttt{gemini-embedding-2} at its maximum output dimensionality of 3,072 to reduce the dimensionality gap with activation representations. We construct two primary representations: \textit{full-prefix embedding} encodes the entire observed prefix, whereas \textit{chunk-averaged embedding} takes the unweighted mean of embeddings of chunks partitioning the prefix at completed event ends. These boundaries come from full-story annotations, although only the observed prefix is embedded. Chunking details and supporting text and length predictors appear in Appendix~\ref{app:text-details}.

Using identical test samples and the previously fixed activation layer, we report activation Macro-AP minus the larger of the two primary text Macro-AP scores. Superiority requires positive differences and Holm-adjusted $p<.05$ against both text predictors; paired bootstrap inference and correction families are specified in Appendix~\ref{app:prediction-inference}. This tests predictive advantage over the specified representations and readouts, not the absence of predictive information from text.

\subsection{Probe-Guided Activation Steering}
\label{sec:activation-steering}

We test whether sustained additive interventions along a probe-derived direction change generated event composition \citep{turner2023activation,rimsky-etal-2024-steering,gao2026raptor}. We target SOCIAL given its strong next-event decodability in OLMo and Gemma, with INTERNAL and WORLD-STATE as supplementary outcomes.

Among the four models, we selected Gemma for the highest activation-probing performance and OLMo for the largest predictive advantage of activations over text-only predictors. We compared them in a pilot with 30 stories per condition at each strength. OLMo exhibited the intended bidirectional control, whereas Gemma did not under the tested settings (Appendix~\ref{app:steering-pilot}). We therefore used OLMo for the main experiment, with 100 seed-matched stories per condition at each strength.

For steering, we select the layer with the highest validation SOCIAL AP at $h=1$, prioritizing next-event information about the intervention target. We then convert its probe coefficients $\mathbf{w}$ to a unit direction in the original activation space:
\begin{equation}
 \mathbf{v}=\frac{\mathbf{w}\oslash\boldsymbol{\sigma}}
 {\|\mathbf{w}\oslash\boldsymbol{\sigma}\|_2},
 \qquad
 \mathbf{x}^{(\ell)}\leftarrow\mathbf{x}^{(\ell)}+\boldsymbol{\delta},
\end{equation}
where $\mathbf{x}^{(\ell)}$ is the hidden state at the selected layer, $\boldsymbol{\sigma}$ contains training feature scales, and $\oslash$ denotes elementwise division. Four conditions use $\boldsymbol{\delta}\in\{\mathbf{0},+\alpha\mathbf{v},-\alpha\mathbf{v},+\alpha\mathbf{r}\}$, with one fixed random unit vector $\mathbf{r}$ orthogonal to $\mathbf{v}$. Strengths are $\{1,2,4,8,16,32\}$ times the training first-event-end projection standard deviation. Stories start from the original prompt, with intervention at its final token and every subsequent decoding step, holding layer and strength fixed.

Stories are annotated without condition metadata using the original event rubric, extended to permit zero-event outputs and record generation-quality diagnostics. The primary outcome is the mean per-story SOCIAL event proportion, assigning zero to zero-event stories and retaining short, truncated, and degenerate outputs. We compare seed-matched conditions using bootstrap intervals and two-sided sign-flip tests, with Holm correction for the two directional comparisons against no steering within each strength. Random-direction comparisons, supplementary outcomes, and implementation details appear in Appendix~\ref{app:steering-details}.

\section{Results}
\begin{figure*}[t]
\centering
\includegraphics[width=\textwidth]{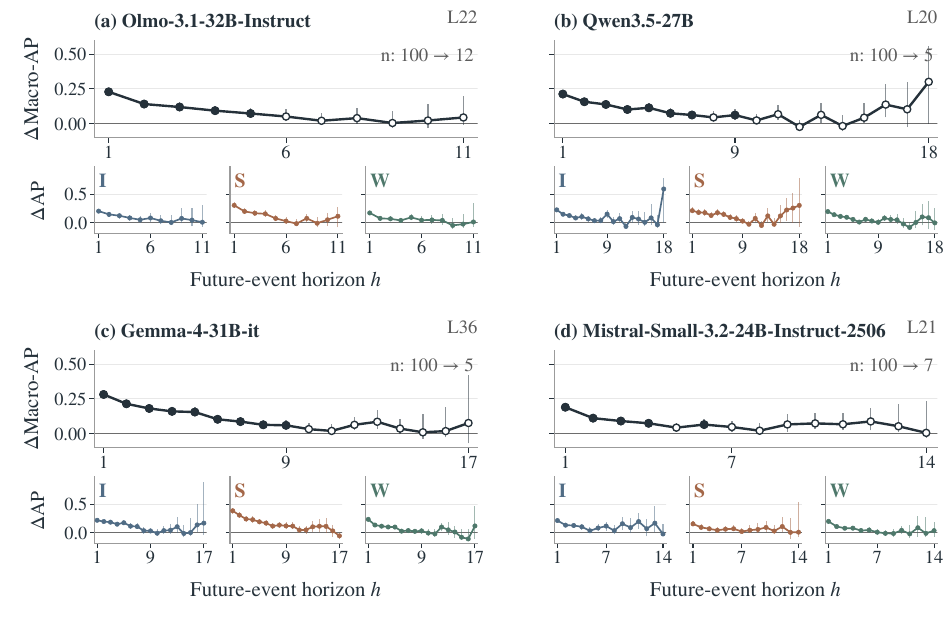}
\caption{Future-event decodability. Each model has a Macro-AP panel above smaller INTERNAL (I), SOCIAL (S), and WORLD-STATE (W) panels. Points show activation AP minus constant-prevalence AP, with pointwise 95\% paired story-bootstrap intervals. Filled Macro markers indicate positive, within-model Holm-significant gains; attribute panels are exploratory. Scales are shared across models within each panel type. L identifies the validation-selected layer; $n$ gives story counts at the first and last horizons. Test anchors are event ends; horizons count not-yet-started events.}
\label{fig:future-event-decodability}
\end{figure*}

\subsection{Event Annotation and Validation}
Across 2,000 stories generated by four models (500 per model), we extracted 23,355 main events, averaging 11.68 events per story (SD = 3.02). The mean number of events per story ranged from 9.08 to 13.48 across models. Averaged across stories, the gap between adjacent annotated event spans was 344 characters (range across models: 220–661), corresponding to 4.6\% of story length (3.2–5.1\% across models). Of all extracted events, 54.5\% were labeled as I, 25.8\% as S, and 58.1\% as W; 35.7\% were assigned multiple state-change attributes. Three human annotators independently assessed the LLM-extracted events and their state-change labels in 30 OLMo stories from the main dataset. A majority accepted 91.9\% of LLM-extracted events as main events. Among 175 events unanimously accepted by the annotators, agreement between LLM labels and human-majority judgments ranged from Cohen's $\kappa = 0.628$ to $0.849$ across the three attributes. Detailed annotation statistics and human validation results are provided in Appendix~\ref{app:event_annotation}.

\subsection{Future-Event Decodability}
\label{sec:results-decodability}

Hidden states predict the attributes of not-yet-started narrative events in all four models (Figure~\ref{fig:future-event-decodability}). At $h=1$, activation Macro-AP is .701 for OLMo, .643 for Qwen, .754 for Gemma, and .656 for Mistral. Gains over constant-prevalence predictions are .229, .213, .282, and .190, respectively, all significant after within-model Holm correction. The attribute panels show positive next-event gains for INTERNAL, SOCIAL, and WORLD-STATE in every model; their pointwise intervals exclude zero, although these attribute-specific analyses are exploratory.

Decodability extends beyond the next event. Positive Holm-significant Macro-AP gains occur at $h=1$--$5$ for OLMo, $h=1$--$7$ and $9$ for Qwen, $h=1$--$9$ for Gemma, and $h=1$--$4$ and $6$ for Mistral. Across the better-supported horizons, gains generally decrease as $h$ increases. However, the contributing anchors and story populations change with $h$, leaving only 5--12 stories at each model's largest eligible horizon. The curves therefore do not identify a sharp boundary beyond which predictive information disappears. Absolute scores, sample counts, and layer and time-resolved analyses appear in Appendix~\ref{app:prediction-results}. These findings support a form of narrative lookahead in which hidden states carry predictive information about the types of several upcoming events.

\subsection{Predictive Advantage over Text-Only Predictors}
\label{sec:results-text}

\begin{figure*}[t]
\centering
\includegraphics[width=\textwidth]{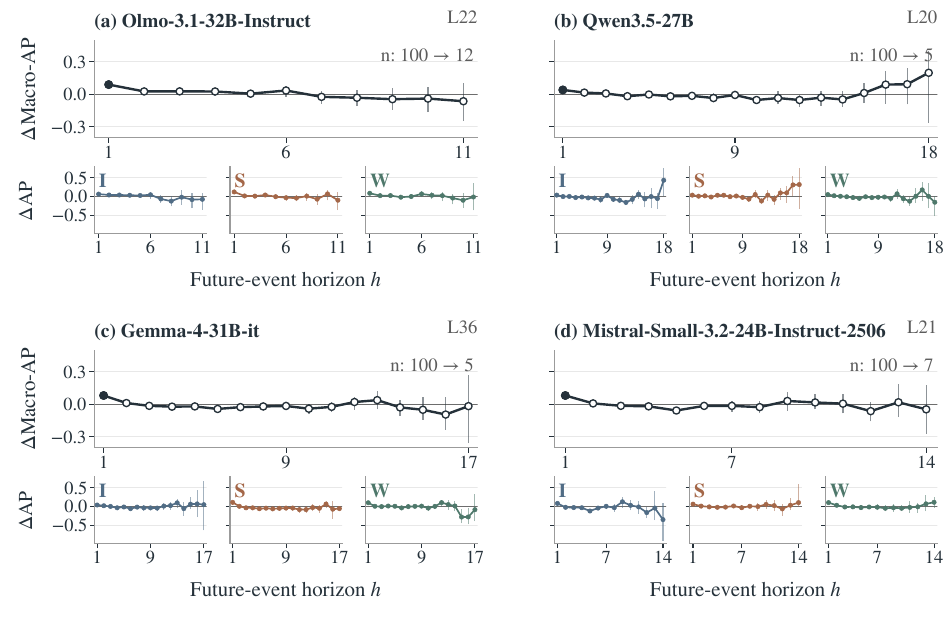}
\caption{Predictive advantage over text-only predictors, using the same model and attribute layout as Figure~\ref{fig:future-event-decodability}. Points show activation AP minus the larger primary text AP for each metric; maxima are recomputed in every paired bootstrap replicate. Whiskers are pointwise 95\% intervals. Filled Macro markers require positive, Holm-significant gains against both predictors in the shared 120-test family. Attribute panels are exploratory: their separate text maxima need not average to the Macro maximum. L and $n$ follow Figure~\ref{fig:future-event-decodability}.}
\label{fig:text-advantage}
\end{figure*}

At $h=1$, activations outperform both primary text predictors in every model (Figure~\ref{fig:text-advantage}). Relative to the larger of full-prefix and chunk-averaged Macro-AP, gains are .090 for OLMo, .041 for Qwen, .081 for Gemma, and .081 for Mistral. All eight pairwise comparisons have Holm-adjusted $p=.012$ within the shared 120-test family. All 12 next-event attribute-level point estimates also favor activations, although the Qwen SOCIAL interval crosses zero.

No $h\geq2$ satisfies the criterion of positive, significant gains against both primary predictors. The evidence for an advantage over these text representations is thus concentrated at the next event, while decodability above prevalence extends further. This comparison concerns the specified representations and linear readouts, and does not establish that the observed text lacks equivalent predictive information. Recent-context embeddings are descriptively stronger than either primary text predictor at $h=1$ for OLMo, Qwen, and Mistral, but remain below activation Macro-AP (Appendix~\ref{app:prediction-results}). The next-event advantage suggests that information about the immediate narrative continuation is more accessible to linear readouts from hidden states than from the two primary text representations.

\subsection{Probe-Guided Activation Steering}
\label{sec:results-steering}

\begin{figure*}[t]
\centering
\includegraphics[width=\textwidth]{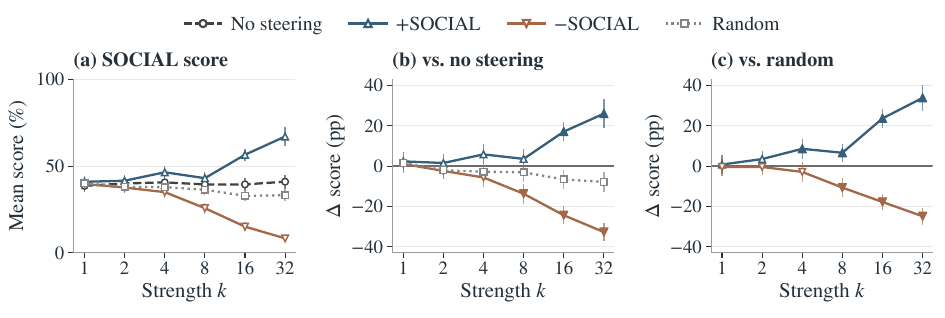}
\caption{OLMo sustained SOCIAL steering ($100$ matched seeds per condition and strength). (a) Mean zero-extended per-story SOCIAL scores. (b,c) Paired differences against no steering and the fixed random direction. Whiskers are pointwise 95\% bootstrap intervals. Filled directional markers in (b,c) indicate within-strength Holm-adjusted $p<.05$ in the respective two-comparison family; random versus no steering is diagnostic. All outputs are retained, including zero-event and flagged outputs. Correction does not extend across strengths.}
\label{fig:olmo-steering}
\end{figure*}

In the 100-seed OLMo experiment, sustained SOCIAL steering changes event composition in the intended directions at $k=16$ and $32$ (Figure~\ref{fig:olmo-steering}a--c). All 30 pilot seeds also appear in this larger sample, so it is not an independent seed replication. At $k=16$, positive steering increases the mean per-story SOCIAL score by 17.13 percentage points (95\% CI [12.45, 21.88]), while negative steering decreases it by 24.36 points ([$-28.89$, $-19.84$]). At $k=32$, the differences are $+26.07$ and $-32.70$ points. All four contrasts have within-strength Holm-adjusted $p<.001$. Both strengths also satisfy the bidirectional criterion against the random direction, which itself reduces SOCIAL scores at these strengths.

Token-normalized SOCIAL event counts change in the same directions as the proportions. The intervention also changes INTERNAL and WORLD-STATE scores. These diagnostics appear in Appendix~\ref{app:steering-results} (Figure~\ref{fig:steering-diagnostics}a,b). Quality costs are visible at the largest strength: positive steering at $k=32$ produces 8/100 disruptive-repetition flags, 5/100 degeneration flags, and 7/100 zero-event outputs (Figure~\ref{fig:steering-diagnostics}c). These potentially overlapping cases remain in the primary analysis. Appendix~\ref{app:steering-results} reports exact contrasts, all-condition diagnostics, and event-count summaries. These findings show that sustained intervention along a direction derived from a next-event SOCIAL probe can alter the subsequently generated events composition in OLMo.

\section{Discussion}
\paragraph{Narrative constraints and immediate prospective accessibility.}
The multi-horizon probing results of Section~\ref{sec:results-decodability} and the activation--text comparisons of Section~\ref{sec:results-text} suggest a distinction between broad narrative constraints and information about the immediately forthcoming event. Character states, relationships, and unresolved conflicts may favor certain kinds of future events without determining which event will occur next. Such constraints could contribute to the multi-horizon decodability observed in Figure~\ref{fig:future-event-decodability}, whereas the consistent activation advantage at $h=1$ in Figure~\ref{fig:text-advantage} suggests that the current generation state makes the attributes of the immediately forthcoming event more readily accessible than the evaluated text representations do. Viewed at the event level, these findings are compatible with narrative lookahead involving broad constraints on later developments alongside more immediate prospective information, without establishing distinct internal representations or a fully specified sequence of future events.

\paragraph{Predictive accessibility and intervention effectiveness need not coincide.} Both Gemma and OLMo show substantial next-event SOCIAL decodability, yet only OLMo exhibits the intended bidirectional changes in SOCIAL event composition under the tested steering protocol. In Gemma, both directions reduced SOCIAL scores at higher strengths, accompanied by widespread degeneration (Appendix~\ref{app:steering-pilot}). Strong decodability alone therefore does not ensure that steering along the corresponding probe direction produces controlled changes in generation. The contrast may reflect differences between models~\citep{silva-etal-2025-steering} as well as aspects of intervention design, such as the choice of intervention layer and how the intervention is parameterized~\citep{sun2025layernavigator, adila2026weight}. A two-model comparison cannot separate these factors. Even where steering is effective, effectiveness is distinct from selectivity~\citep{stoehr-etal-2024-activation}: OLMo's changes extend to INTERNAL and WORLD-STATE attributes, and stronger interventions introduce generation-quality costs. While localized activation patching can test whether decodable future information causally contributes to generation~\citep{ma2026where}, our sustained intervention does not establish such a role for the probe-derived direction at particular event boundaries or during unperturbed generation. Nevertheless, the OLMo results demonstrate that intervening along a direction derived from next-event prediction can causally reshape the composition of subsequently generated narrative events.

\paragraph{Event-level lookahead and narrative planning.} The event level offers a way to investigate narrative planning without assuming that a model commits to a particular story trajectory in advance. Narrative generation unfolds through successive developments that can alter the possibilities available to later events. Whereas prior research has examined advance information about response-level attributes \citep{dong2025emergent} or used event representations and explicit plans to guide story generation \citep{Martin_2018,razumovskaia-etal-2024-little}, our approach examines narrative changes as targets of prospective prediction rather than as externally supplied plans. Our plot-independent attributes identify kinds of development without specifying particular trajectories. This perspective shifts the question from whether a model has already determined what will happen to whether its current state carries information about what kinds of narrative developments may follow. Such information is compatible with planning through constraints on plausible continuations rather than a fixed sequence of events, but need not reflect a dedicated planning computation, as suggested by the distinction between pre-caching and breadcrumbs \citep{wu2024languagemodelsplanahead}. Nor can a single realized continuation establish how a model represents alternative possible futures. Determining whether this lookahead constitutes an operative planning process will require tracing how future-event information evolves across generation and testing its localized influence on later developments.

\section{Conclusion}
We investigate narrative lookahead through salient, state-changing events, using event-based horizons and plot-independent labels to compare prospective information across stories. Across four open-weight models, future event types are decodable several events ahead, yet hidden-state probes reliably outperform both primary text-only predictors only for the next event. In exploratory steering with OLMo, sustained intervention along a direction derived from the next-event SOCIAL probe shifts the proportion of SOCIAL events in both intended directions. By making plot-independent event types both targets of future prediction and measures of subsequent narrative composition, our approach connects prospective information in hidden states to intervention effects on generated narratives. This event-level framework provides a basis for investigating whether and how such information guides narrative generation without a predetermined plot.

\section*{Limitations}
All stories were generated by four 24–32B instruction-tuned models using a single minimal English prompt and one sampling temperature. This setting supports comparisons across models under a common generation task, while leaving open whether the findings generalize to different prompts, genres, model scales, or human-authored narratives.

Events and their labels come from a single proprietary LLM, validated against human judgments only on OLMo-generated stories. Because event status and locus are interpretive, human agreement defines a range rather than a fixed standard, and LLM labels fell within the human range. Event types from steered outputs were annotated by the same LLM without separate human validation. 

The text-only comparisons use matched prefixes and linear readouts, and stronger text-based predictors remain to be evaluated. The main steering experiment focused on SOCIAL events in OLMo, intervening at one selected layer throughout generation. These results demonstrate changes in aggregate event composition under the tested intervention protocol, but do not establish that a hidden state at a particular event boundary causally determines the next event. Higher steering strengths also produced some generation-quality costs.

\bibliography{custom}

\clearpage
\appendix
\PaperFloatBarrier
\raggedbottom

\section{Story Generation Details}
\label{app:generation-details}

\subsection{Checkpoints, Prompts, and Sampling}

The four model identifiers are 
\texttt{allenai/Olmo-3.1-32B-Instruct}, \texttt{Qwen/Qwen3.5-27B}, \texttt{google/gemma-4-31B-it}, and \texttt{mistralai/}\allowbreak\texttt{Mistral-Small-3.2-}\allowbreak\texttt{24B-Instruct-2506}.

We applied each tokenizer's chat template with \texttt{add\_generation\_prompt=True}. The sole user message was  \texttt{Write a long story}. We supplied no additional system message; OLMo's template inserted its default system message. Qwen and Gemma use their non-thinking modes; Gemma explicitly passes \texttt{enable\_thinking=False}.

We generated 30 pilot stories for each combination of OLMo and Qwen with temperatures ${0.6,0.9,1.1,1.3}$. We examined abstract-event distributions, semantic diversity, and signs of degeneration, and selected 0.9 as a setting with relatively high semantic diversity across stories and no evident degeneration.

Generation proceeded one token at a time with KV caching and frozen model weights. Logits were cast to FP32, divided by temperature, and nucleus-filtered before multinomial sampling. The top-$p$ implementation retained the first token crossing cumulative probability 0.95. We applied no top-$k$, repetition, or minimum-length logits processor. EOS was not suppressed before 500 generated tokens; an earlier EOS caused the attempt to be rejected. Token counts exclude the prompt and include EOS. An EOS at generated token 4,000 was accepted, whereas a non-EOS token at that position caused immediate rejection. Stop-token IDs were read from each model's generation configuration, falling back to the tokenizer when necessary. Eight Qwen attempts reached the 4,000-token limit without EOS, and two Mistral attempts ended before 500 tokens; all were replaced. OLMo and Gemma had no length-based rejections.

All four models were loaded with Transformers 5.17.0, unquantized weights, and BF16 when supported (FP16 otherwise). Generation used a fixed seed schedule, assigning a new seed to each attempt within a model.

\subsection{Boundary Candidates and Saved States}
\label{app:boundary-details}

Generation and annotation share a deterministic rule for identifying candidate sentence boundaries. It defines candidate units rather than a linguistically exact sentence segmentation. Three boundary types are recognized: (i) runs of sentence-final punctuation, excluding common abbreviations, initials, and dotted abbreviations; (ii) ellipses followed by a closing symbol or not continued by a lowercase word on the same line; and (iii) nonempty line endings not already represented by a punctuation or ellipsis boundary. Closing quotes, brackets, and Markdown emphasis markers are skipped when identifying a boundary core. The rule can include headings and utterances before dialogue tags. Such candidates need not be narrative events.

We decode the complete generated prefix with special tokens skipped and tokenization-space cleanup disabled. Trailing incomplete UTF-8 replacement characters are deferred until resolved. A candidate's state is saved after processing the token that first completes its boundary character, even when later text is needed to confirm it. Multiple boundaries in one token share a state. The state record includes the boundary's character end, generating token index, exclusive visible-text cutoff, confirmation step, and flags for shifted states or tokens containing subsequent narrative content. These records and the saved token IDs support exact downstream event alignment.

Activations are the model's returned \texttt{hidden\_states}, excluding the embedding entry, with shape [saved time, layer, hidden dimension]. They are not uniformly described as pre-normalization residual streams: the final returned layer may already include final normalization. We retain raw generated text, including titles or framing text, and do not apply a separate semantic non-narrative-output classifier. Event selection subsequently determines which spans enter the narrative representation.

\PaperFloatBarrier
\section{Narrative Event Annotation Details}
\label{app:annotation-details}

\subsection{API Protocol and Validation}

We annotated stories using the Responses API with \texttt{gpt-6-sol}, structured Pydantic output, and \texttt{max\_output\_tokens=16000}. Sampling and reasoning parameters were left unspecified. Each request contained the fixed annotation rubric and complete story, without story identifiers or generation-model metadata. 

Responses were validated for nonempty event lists, sequential event numbering, verbatim spans, valid binary domain labels (at least one positive), and Attempted/Realized status. Invalid responses were retried up to three times with error-specific feedback while retaining the original rubric. Exhausted attempts were recorded as failures rather than valid annotations. Valid annotations were reused across reruns.

\subsection{Span Expansion and Downstream Alignment}

After verbatim validation, a span beginning or ending inside a candidate sentence unit is expanded to the smallest enclosing units using the generation boundary rule. This affected 28 event spans (OLMo: 22; Qwen: 4; Gemma: 1; Mistral: 1). Original spans are retained in \texttt{original\_span}, alongside expansion flags. Labels and transition status are retained without another annotation call. Closers or whitespace alone are handled as boundary suffixes rather than additional narrative content. This is a deterministic alignment operation, not a new semantic judgment of the enlarged span; added context can extend beyond the originally selected event.

The annotation validator checks increasing event numbers but does not verify textual order, disjointness, or unique span occurrences. Post-annotation anchor checks identified 16 stories with event numbering inconsistent with narrative order (OLMo: 2; Qwen: 11; Gemma: 1; Mistral: 2); these stories were regenerated and reannotated. Expansion can create overlapping or identical spans. During prediction-data construction, spans must occur uniquely in the story, are sorted by their textual starts, and must have distinct start positions. Ambiguous repeated spans or identical starts cause an error requiring annotation review; they are not silently assigned arbitrary event positions. Overlap with distinct starts is allowed, and already-started events are excluded from future targets. Exact boundary and token-cutoff failures remove individual anchors according to Appendix~\ref{app:prediction-details}.

\PaperFloatBarrier
\section{Narrative Event Annotation Results and Validations}
\label{app:event_annotation}

\subsection{Descriptive Statistics}
\label{app:descriptive statistics}
Table \ref{tab:event_annotation_stats} summarizes the event annotations across the four models. The mean number of events per story ranged from 9.08 (OLMo) to 13.48 (Qwen), with Mistral and Gemma averaging 10.89 and 13.26, respectively. Domain distributions also varied across models: INTERNAL and WORLD-STATE labels were relatively prevalent across all four, whereas SOCIAL labels showed greater variation, ranging from 19.2\% (Qwen) to 37.5\% (OLMo). 
\begin{table}[!htbp]
\centering
\small
\setlength{\tabcolsep}{3pt}
\begin{tabular}{lrrrr}
\toprule
 & OLMo & Qwen & Gemma & Mistral \\
\midrule
Events & 4,541 & 6,741 & 6,630 & 5,443 \\
Events/story (mean) & 9.08 & 13.48 & 13.26 & 10.89 \\
Events/story (SD) & 1.96 & 2.72 & 2.21 & 2.71 \\
Events/story (range) & 4--17 & 6--22 & 7--21 & 5--20 \\
INTERNAL (\%) & 52.6 & 51.3 & 59.7 & 53.7 \\
SOCIAL (\%) & 37.5 & 19.2 & 26.4 & 23.4 \\
WORLD-STATE (\%) & 53.4 & 59.5 & 56.1 & 62.9 \\
Multi-domain (\%) & 40.1 & 28.2 & 38.7 & 37.5 \\
Span length (words) & 32.7 & 33.4 & 36.7 & 25.3 \\
\bottomrule
\end{tabular}
\caption{Event annotation statistics across four models (500 stories per model). Domain labels are not mutually exclusive. Multi-domain denotes events assigned to two or more domains.}
\label{tab:event_annotation_stats}
\end{table}

\subsection{Annotation Stability}
To assess run-to-run stability, we ran the annotation pipeline three times
on 30 OLMo stories randomly sampled from the 500 used in the main experiment.
Mean pairwise exact-span F1 was .708, and overlap-based F1 was .837,
using order-preserving one-to-one matching at character-span IoU~$\geq$~.5.
On the 680 matched event pairs pooled across run pairs, Cohen's $\kappa$
was .834 for INTERNAL, .895 for SOCIAL, and .894 for
WORLD-STATE.

\subsection{Human Validation}
\label{app:human-validation}
Human validation assessed the interpretability and discriminative validity of our operational definition of main narrative events.
We selected 30 stories generated by OLMo-3.1-32B-Instruct from the main dataset. The validation set comprised 270 LLM-extracted event spans and 138 control spans drawn from story text outside the extracted events. Controls were manually classified as easy (n = 61; descriptive or background material lacking a salient narrative event) or hard (n = 77; actions or minor changes not clearly qualifying as main events). This design tested both the rejection of relatively unambiguous non-events and the boundary between local narrative activity and main-event salience.

Three human annotators independently evaluated all 408 candidates. Each complete story was presented in its original narrative context, with one candidate span highlighted at a time.  LLM-extracted events and controls were intermixed, with their source and control difficulty concealed. For each candidate, human annotators judged main-event status (Yes/No) and assigned binary INTERNAL, SOCIAL, and WORLD-STATE labels using a rubric with the same annotation criteria as the LLM annotator, supplemented with examples to facilitate human understanding. Main-event acceptance required at least two positive judgments; domain-label agreement was evaluated on LLM-extracted events unanimously accepted by all three annotators.
As shown in Table \ref{tab:event_validation}, 91.9\% of LLM-extracted events were accepted by a human majority, compared with 4.9\% of easy controls and 28.6\% of hard controls. Agreement on domain labels for the 175 unanimously accepted events is reported in Table \ref{tab:domain_validation}.

\begin{table}[!htbp]
\centering
\small
\setlength{\tabcolsep}{4pt}
\begin{tabular}{lrrr}
\toprule
Candidate group & $N$ & Majority (\%) & Pooled (\%) \\
\midrule
LLM events & 270 & 91.9 & 85.2 \\
Easy controls & 61 & 4.9 & 7.1 \\
Hard controls & 77 & 28.6 & 31.2 \\
\bottomrule
\end{tabular}
\caption{Human validation of main-event extraction
across 408 candidates from 30 OLMo-generated stories.
Majority denotes acceptance by at least two of three
annotators; pooled denotes the proportion of positive
judgments across all annotators.
Human inter-annotator agreement was
Fleiss' $\kappa = 0.543$.}
\label{tab:event_validation}
\end{table}

\begin{table}[!htbp]
\centering
\small
\setlength{\tabcolsep}{5pt}
\begin{tabular}{lrrrr}
\toprule
 & \multicolumn{3}{c}{$\kappa$} & \\
\cmidrule(lr){2-4}
Domain & H--H & L--Ind. & L--Maj. & Pos. F1 \\
\midrule
INTERNAL & 0.579 & 0.585 & 0.628 & 0.809 \\
SOCIAL   & 0.552 & 0.689 & 0.764 & 0.839 \\
WORLD-STATE    & 0.590 & 0.706 & 0.849 & 0.934 \\
\bottomrule
\end{tabular}
\caption{Agreement on narrative domain labels
for 175 LLM-extracted events unanimously
accepted as main events by three human annotators.
H--H: human Fleiss' $\kappa$;
L--Ind.: mean LLM--individual Cohen's $\kappa$;
L--Maj.: LLM--human-majority Cohen's $\kappa$;
Pos. F1: positive-class F1 against human-majority
labels. Each domain is evaluated independently.}
\label{tab:domain_validation}
\end{table}

\PaperFloatBarrier
\section{Prediction Analysis Details}
\label{app:prediction-details}

\subsection{Anchors, Targets, and Splits}

We use the shared rule for identifying candidate sentence boundaries described in Appendix~\ref{app:boundary-details}. An event-end anchor uses the saved activation at the token completing the event's final boundary core. Trailing closing symbols or whitespace need not have been consumed. We exclude anchors without an exact saved boundary, shifted activations, and tokens that already include subsequent narrative text or the beginning of another event. We do not substitute nearby activations. Events sharing an end token yield one anchor, indexed by their largest event ordinal for time-resolved analyses.

Future targets are ordered events whose character start is at or beyond the activation's exclusive character cutoff. Thus, with nonoverlapping spans, horizon $h$ after event $E_i$ corresponds to $E_{i+h}$. Already-started overlapping events are skipped. An event remains a possible target even if its own end cannot be used as an anchor. Both Attempted and Realized events contribute their original multi-label attributes.

For each model, sorted story identifiers are permuted once with NumPy's default random-number generator using the same fixed random seed, then split 60/20/20. This is a fixed story-level holdout split. The prompt-end anchor $E_0$ is retained in training and validation for regularization selection; representative-layer selection and all reported test analyses use event-end anchors only.

\subsection{Probe Estimation and Layer Selection}

For a fixed layer and horizon, each feature is standardized using the unweighted training-sample mean and population standard deviation (\texttt{ddof=0}); constant features receive scale one. These statistics are reused without modification on validation and test data. Writing $z_i=\mathbf{w}^{\top}\widetilde{\mathbf{x}}_i+b$, we minimize
\begin{equation}
 \mathcal{L}(\mathbf{w},b)=
 \sum_i\frac{\log(1+e^{z_i})-y_i z_i}{S n_{s(i)}}
 +\frac{\lambda}{2}\|\mathbf{w}\|_2^2,
\end{equation}
where $S$ is the number of represented training stories and $n_{s(i)}$ their horizon-specific sample count. The intercept is unpenalized. We apply no additional class weighting or oversampling. Story weighting is recomputed within each training, validation, or evaluation subset.

We search $\lambda\in\{0.01,0.1,1,10,100\}$ independently for each binary probe, maximizing weighted validation AP; exact ties favor larger $\lambda$. Optimization uses SciPy L-BFGS-B without bounds, zero coefficient initialization, and an intercept initialized to the logit of weighted training prevalence. Gradient and function tolerances are $10^{-6}$ and $10^{-12}$, respectively, with up to 50 line-search steps. The iteration limits are 5,000 for activation probes and 10,000 for text predictors; the function-evaluation limit is five times the iteration limit. Unconverged fits are not accepted. Models are not refitted on combined training and validation data.

A binary probe is unavailable if its training or validation targets lack either class. For representative-layer selection, we use horizons with both classes for all three attributes in the event-end training and validation subsets and valid probes at every layer. This common set is defined within each model, independently of test labels. We average validation Macro-AP equally across these horizons and break exact layer ties toward the earlier layer. Table~\ref{tab:probe-settings} records the resulting choices. The input embedding layer is excluded from layer numbering.

\begin{table}[t]
\centering
\small
\begin{tabular}{lrrr}
\toprule
Model & Layers & Selected & Horizons \\
\midrule
OLMo & 64 & 22 & 1--11 \\
Qwen & 64 & 20 & 1--18 \\
Gemma & 60 & 36 & 1--17 \\
Mistral & 40 & 21 & 1--14 \\
\bottomrule
\end{tabular}
\caption{Validation-selected prediction layers (one-based) and eligible horizons. The same selected layers are used in the observable-text comparisons.}
\label{tab:probe-settings}
\end{table}

\subsection{Observable-Text Features}
\label{app:text-details}

Each input is the original generated prefix ending at the saved activation cutoff. Embedding requests prepend the exact instruction \texttt{task: classification | query: } and request 3,072 output dimensions from \texttt{gemini-embedding-2}, with automatic truncation disabled. Embeddings are cached by input text and embedding configuration. Empty prefixes at $E_0$ receive zero vectors without an API request.

For the chunk-averaged representation, cuts occur at annotated event ends whose boundary cores have already been observed, clipped to the current cutoff. The chunks partition the entire prefix without overlap or omission, including intervening text; whitespace-only tails are attached to the preceding chunk. We take the unweighted arithmetic mean of chunk embeddings without subsequent vector normalization. This is annotation-informed segmentation: boundaries come from whole-story annotations, although no future text or event-type labels enter the embedding inputs. It is not an independently online event-boundary detector.

The recent-context representation embeds the last three observed sentence units, or the entire prefix if fewer are available, retaining any visible trailing suffix. The length representation contains two features: the number of observed sentence boundaries and the number of generated tokens. TF--IDF uses lowercased word unigrams and bigrams, tokens matching \texttt{(?u)\textbackslash b\textbackslash w\textbackslash w+\textbackslash b}, raw term frequencies, smoothed inverse document frequencies, and document-level $\ell_2$ normalization. There is no stop-word removal or vocabulary-size cap. Vocabulary and IDF are fitted separately for each horizon using unique, nonempty training prefixes. Dense and length features receive training-only standardization; TF--IDF features receive no additional standardization. All five predictors use the probe objective and selection procedure above.

\subsection{Metrics and Statistical Inference}
\label{app:prediction-inference}

AP is computed once over pooled predictions with weights $a_i=1/(S n_{s(i)})$, not averaged across stories. Macro-AP is the arithmetic mean of all three attribute APs. The prevalence baseline assigns the horizon- and attribute-specific story-balanced training prevalence as a constant prediction. Its test AP equals weighted test prevalence as a consequence of constant-score AP, without fitting on test labels. Attribute-specific AP is a supplementary metric.

For paired bootstrap inference, we sample the $S$ eligible test stories with replacement and use identical story multiplicities for all compared predictors and attributes. If story $s$ appears $m_s$ times, its samples receive weights $m_s/(S n_s)$. We recompute pooled AP and contrasts in every replicate, keeping fitted probes, regularization strengths, and selected layers fixed. Intervals are the 2.5th and 97.5th percentiles and quantify test-story sampling uncertainty conditional on these fitted models. For the maximum-text comparison, we recompute the maximum of the two text Macro-APs in every replicate.

Pairwise tests use the two-sided centered-bootstrap approximation
\begin{equation}
 p=\frac{1+\sum_{b=1}^{B}\mathbb{1}
 \{|\Delta_b^*-\widehat\Delta|\geq|\widehat\Delta|\}}
 {B+1}.
\end{equation}
Confidence intervals are pointwise rather than simultaneous, and need not agree exactly with these tests. Bootstrap seeds are deterministically derived from a fixed base seed for each run and subset and recorded by the implementation.

For activation versus prevalence, Holm correction is applied separately within each model: all eligible-horizon Macro-AP tests form the primary family, and all nonempty anchor-by-horizon Macro-AP tests form a separate secondary family. The replicate count is $B=\max(10{,}000,100M)$ for a family of size $M$.

For activation versus text, the two primary contrasts are activation versus full-prefix and activation versus chunk-averaged embeddings. The horizon-level Holm family combines both contrasts across all four models and their eligible horizons; the time-resolved family separately combines both contrasts across all models and nonempty anchor-by-horizon cells. These analyses use 10,000 replicates. A positive maximum-text advantage is marked significant only when both pairwise differences are positive and both adjusted $p$-values are below .05. Attribute-specific and supporting-predictor comparisons remain exploratory.

Macro-level inference requires valid predictions and both target classes for all three attributes, with at least two eligible test stories. Unavailable planned comparisons remain in their correction family with $p=1$. For descriptive AP, all-negative targets receive 0 and all-positive targets receive 1. Single-class bootstrap replicates are retained, and their frequency is recorded. Structurally empty time-by-horizon cells are omitted.

\PaperFloatBarrier
\section{Activation Steering Details}
\label{app:steering-details}

\subsection{Direction, Calibration, and Generation}

Steering layer selection maximizes the saved validation AP of the horizon-one SOCIAL probe, including the original prompt-end validation samples, with ties favoring the earlier layer. This criterion differs from the Macro-AP criterion used for prediction. The selected layers are 27 for OLMo and 41 for Gemma. The unit direction is the normalized vector of probe coefficients divided elementwise by the training feature scales. We retain the original training-only probe; our story-weighted AP selection and fixed strength sweep are study-specific adaptations of the probe-derived direction approach of \citet{gao2026raptor}.

The calibration set contains training stories with exactly one saved anchor belonging only to $E_1$, no ongoing overlapping event, and $E_2$ as the next unstarted target. We project their raw activations onto the SOCIAL direction and compute the population standard deviation $\sigma_{\mathrm{proj}}$. Both models have 300 eligible calibration stories. The recorded scales are 1.6887418371 for OLMo and 0.0450376721 for Gemma. Intervention magnitude is $\alpha=k\sigma_{\mathrm{proj}}$, for $k\in\{1,2,4,8,16,32\}$. A Gaussian random vector, drawn once using a fixed random seed, is projected off the SOCIAL direction and normalized. This single random direction is held fixed across strengths and receives the same magnitude as the directional interventions.

We generate from the user message \texttt{Write a long story}, using the model's saved revision and chat template with \texttt{add\_generation\_prompt=True}. Gemma additionally uses \texttt{enable\_thinking=False}. Sampling uses temperature 0.9, top-$p=0.95$, no top-$k$ filtering, and a limit of 4,000 new tokens including EOS. We impose no minimum length and do not suppress EOS. Model weights remain frozen and unquantized, using BF16 and scaled dot-product attention with Transformers 5.17.0. The hook is checked against the hidden-state tensor used for probe training; the final-layer case checks whether the matching tensor is before or after the final normalization.

The hook modifies only the current prediction position, first at the last prompt token and then once per cached decoding step; past cache entries are not re-edited. Each seed initializes a separate CUDA random-number generator and is reused across the four conditions and all strengths within a phase. The OLMo seed schedule deterministically derives phase- and story-specific seeds from a fixed base seed. The no-steering condition is generated within each strength run. Full seed lists, model revision hashes, sampling settings, directions, and hook checks are recorded in run manifests.

\subsection{Model-Selection Pilot}
\label{app:steering-pilot}

The pilot used 30 seed-matched stories per condition at each of six strengths for both Gemma-4-31B-it and OLMo-3.1-32B-Instruct (720 outputs per model); all planned outputs were generated and annotated. Table~\ref{tab:steering-pilot} reports the SOCIAL contrasts and quality diagnostics. OLMo met the bidirectional-support criterion at $k=16$ and $32$: positive steering increased the score and negative steering decreased it, with both within-strength Holm-adjusted $p<.05$. Gemma met this criterion at no tested strength. At $k=8$, both Gemma directions decreased the score, with narrative degeneration flagged in 22/30 positive and 29/30 negative outputs; at $k=16$ and $32$, all outputs in both directions were flagged as degenerate and had no annotated events. OLMo's degeneration count never exceeded 2/30 per condition. These observations motivated selecting OLMo for the main experiment and concern only the tested steering protocol.

\paragraph{Relation to the 100-seed experiment.}
The saved OLMo seed lists contain 30 unique pilot seeds and 100 unique main-experiment seeds, with all 30 pilot seeds also present in the larger schedule. The 100-seed analysis is therefore a larger-sample evaluation following pilot-based model selection, not an independent seed replication. We analyze its 100 seeds once and do not pool the pilot as 30 additional independent observations. The overlap concerns generation seeds; matching seed identifiers alone does not establish byte-identical outputs across separately executed runs.

\begin{table*}[t]
\centering
\small
\setlength{\tabcolsep}{4pt}
\begin{tabular}{lrllccc}
\toprule
Model & $k$ & $\Delta_{+}$ [95\% CI] & $\Delta_{-}$ [95\% CI] & R ($+/-$) & D ($+/-$) & Z ($+/-$) \\
\midrule
Gemma & 1 & $-1.14$ [-7.49, 5.36] & $-1.52$ [-6.90, 3.70] & 0/0 & 0/0 & 0/0 \\
 & 2 & $+2.83$ [-2.67, 8.14] & $-0.39$ [-6.64, 6.09] & 0/0 & 0/0 & 0/0 \\
 & 4 & $+0.47$ [-4.60, 5.51] & $+1.60$ [-4.92, 7.90] & 1/0 & 0/0 & 0/0 \\
 & 8 & $-11.68^{*}$ [-17.38, -6.01] & $-13.88^{*}$ [-21.74, -5.10] & 23/29 & 22/29 & 0/7 \\
 & 16 & $-23.67^{*}$ [-27.89, -19.44] & $-23.67^{*}$ [-27.89, -19.44] & 25/30 & 30/30 & 30/30 \\
 & 32 & $-26.46^{*}$ [-30.44, -22.10] & $-26.46^{*}$ [-30.44, -22.10] & 30/29 & 30/30 & 30/30 \\
\midrule
OLMo & 1 & $+0.55$ [-8.92, 10.09] & $-3.30$ [-11.24, 4.30] & 0/0 & 0/0 & 0/0 \\
 & 2 & $+0.49$ [-8.40, 8.87] & $-0.85$ [-7.33, 5.49] & 0/0 & 0/0 & 0/0 \\
 & 4 & $+12.31^{*}$ [2.83, 21.45] & $-4.81$ [-13.13, 3.59] & 0/0 & 0/0 & 0/0 \\
 & 8 & $+8.17$ [-0.82, 17.35] & $-9.41$ [-17.95, -0.80] & 0/0 & 0/0 & 0/0 \\
 & 16 & $+12.84^{*}$ [3.54, 22.41] & $-26.30^{*}$ [-35.58, -17.58] & 0/0 & 0/1 & 0/1 \\
 & 32 & $+31.06^{*}$ [17.66, 43.42] & $-32.55^{*}$ [-40.64, -24.83] & 4/0 & 2/0 & 1/0 \\
\bottomrule
\end{tabular}
\caption{Model-selection pilot ($n=30$ per condition and strength). $\Delta_{+}$ and $\Delta_{-}$ are paired differences in mean per-story SOCIAL score versus no steering, in percentage points. CIs are pointwise paired-bootstrap intervals; $^{*}$ denotes a two-sided sign-flip test with Holm-adjusted $p<.05$ across the two directional comparisons within a strength, without correction across strengths. R, D, and Z count outputs with annotator-flagged disruptive repetition, narrative degeneration, and zero events, respectively, separately for positive/negative steering. All outputs are retained; zero-event outputs receive score zero. No-steering and random-direction controls had no degenerate or zero-event outputs.}
\label{tab:steering-pilot}
\end{table*}

\subsection{Annotation and Outcomes}

Annotation follows the API protocol and span-alignment procedure in Appendix~\ref{app:annotation-details}, retaining the original event rubric, multi-label attributes, and Attempted/Realized metadata. The annotator receives no steering condition, strength, or seed; task identifiers are randomized using a fixed seed. The rubric is extended only to permit empty event lists and return Boolean flags for disruptive repetition and narrative degeneration. These flags do not alter event selection or exclude stories. Each pass allows up to four API attempts, including the initial call, with error-specific feedback. Failed records remain incomplete rather than counting as zero-event stories; recovery passes revisit missing or invalid records while preserving the seed set. The exact additional instructions and extended output format used for OLMo steering are provided in Appendix~\ref{app:steering-prompt-extension}.

For story $s$ under condition $a$, let $N_{s,a}$ be its annotated event count. For attribute $c$, the analyzed score is
\begin{equation}
 R_{s,a,c}=\begin{cases}
 N_{s,a}^{-1}\sum_{j=1}^{N_{s,a}}y_{s,a,j,c}, & N_{s,a}>0,\\
 0, & N_{s,a}=0.
 \end{cases}
\end{equation}
We average this score equally across seeds, with SOCIAL as primary. The zero extension is an explicit outcome convention, not a claim that an empty story has a defined event proportion. We also report zero-event rates, event counts, label counts per 1,000 generated tokens, length, EOS before 32 content tokens, length-limit hits, annotation quality flags, and repeated token 4-gram fractions. Ratios conditional on eventful outputs are descriptive. Multi-label attributes need not sum to one.

\subsection{Paired Inference}

At each strength, all four outputs for each seed must have valid annotations before final paired inference. Mean differences are reported in percentage points, with 95\% percentile intervals from 10,000 bootstrap resamples of complete seed blocks using a fixed random seed. Two-sided paired sign-flip tests use the absolute mean difference. We enumerate all sign assignments when at most 18 paired differences are nonzero; otherwise, we use 100,000 random sign assignments generated with a separate fixed random seed and a plus-one correction. The test assumes sign exchangeability of paired differences under the null.

For SOCIAL, the two comparisons $+\mathrm{SOCIAL}$ versus no steering and $-\mathrm{SOCIAL}$ versus no steering form one Holm family. The two corresponding comparisons against the random direction form a separate supporting family. Bidirectional support requires the positive intervention to increase the score and the negative intervention to decrease it, with both adjusted $p<.05$. INTERNAL and WORLD-STATE each use the same outcomes and paired procedure; their four comparisons against no steering form one exploratory family, and their four comparisons against random form another. Random versus no-steering tests are unadjusted diagnostics. Corrections are applied within each strength, not jointly across strengths; the strength sweep is therefore a descriptive sensitivity analysis rather than a familywise test over the full sweep.

\PaperFloatBarrier
\section{Additional Prediction Results}
\label{app:prediction-results}

\subsection{Corpus Composition and Sample Support}

Corpus-level attribute frequencies (Table~\ref{tab:event_annotation_stats}) differ from the story-balanced test prevalence underlying the constant-score AP baseline. Because each model is evaluated on its own generated stories, cross-model AP differences also reflect differences in event composition and label distributions.

The final prediction caches contain 4,041, 6,241, 6,130, and 4,943 event-end anchors for OLMo, Qwen, Gemma, and Mistral, respectively, after omitting anchors with no future event. The final exclusion summaries record 500 such omissions per model and no additional exclusions for the other recorded reasons. Tables~\ref{tab:horizons-olmo}--\ref{tab:horizons-mistral} give horizon-specific test sample counts. A story can contribute multiple anchors, but its total weight within a horizon is the same as that of any other eligible story.

\subsection{Absolute Scores and Attribute-Specific Results}

Table~\ref{tab:next-event-predictors} reports absolute $h=1$ Macro-AP for every predictor; Tables~\ref{tab:horizons-olmo}--\ref{tab:horizons-mistral} provide the corresponding contrasts, confidence intervals, adjusted $p$-values, and sample counts.

\begin{figure*}[tp]
\centering
\includegraphics[width=\textwidth]{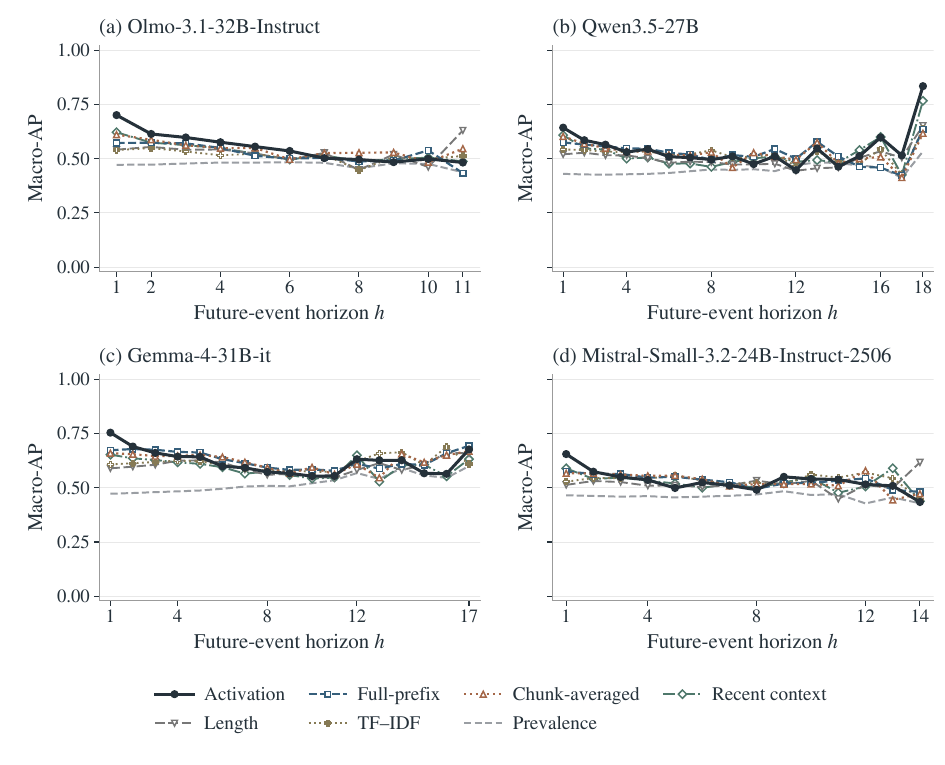}
\caption{Absolute story-balanced Macro-AP for all predictors across future-event horizons at the fixed validation-selected prediction layers. All predictors use matched test samples. Curves are descriptive point estimates. Only training/validation-eligible horizons are shown; single-class targets follow the descriptive conventions in Appendix~\ref{app:prediction-inference}.}
\label{fig:all-predictors-ap}
\end{figure*}

Full-prefix and chunk-averaged embeddings define the primary text comparison. The recent-context, TF--IDF, and length predictors are supporting analyses and are not part of its superiority criterion. Figure~\ref{fig:all-predictors-ap} shows AP for all predictors across horizons.

Table~\ref{tab:attribute-next-event} reports next-event attribute-specific AP and contrasts, complementing the INTERNAL, SOCIAL, and WORLD-STATE panels in Figures~\ref{fig:future-event-decodability} and~\ref{fig:text-advantage}. At $h=1$, all 12 activation-minus-prevalence intervals exclude zero. The largest SOCIAL gains occur for OLMo (.309) and Gemma (.389), while Qwen and Mistral have lower absolute SOCIAL AP (.400 and .402). These descriptive cross-model differences depend on the generated corpora and label distributions. Attribute-specific intervals are pointwise and unadjusted.

Against the larger primary text AP for each attribute, all 12 next-event point estimates are positive. Eleven intervals exclude zero; Qwen SOCIAL has a gain of .036 with interval [$-.015$, $.073$]. These are exploratory comparisons rather than a second family of primary tests. Each attribute uses its own maximum over the full-prefix and chunk-averaged scores in every bootstrap replicate. Consequently, averaging the three displayed attribute gains need not recover the Macro gain, which uses the maximum of the two Macro-APs.

\begin{table*}[tp]
\centering
\small
\setlength{\tabcolsep}{4.2pt}
\begin{tabular}{lrrrrrrr}
\toprule
Model & Activation & Full-prefix & Chunk-avg. & Recent & TF--IDF & Length & Prevalence \\
\midrule
OLMo & \textbf{.701} & .572 & .611 & .622 & .539 & .541 & .472 \\
Qwen & \textbf{.643} & .575 & .602 & .609 & .542 & .520 & .430 \\
Gemma & \textbf{.754} & .673 & .660 & .653 & .608 & .590 & .473 \\
Mistral & \textbf{.656} & .575 & .567 & .590 & .528 & .513 & .466 \\
\bottomrule
\end{tabular}
\caption{Next-event ($h=1$) story-balanced Macro-AP. Full-prefix and chunk-averaged embeddings are the two primary text predictors; Recent, TF--IDF, and Length are supporting predictors. Bold identifies activation scores for readability, not additional significance tests. All rows use 100 test stories.}
\label{tab:next-event-predictors}
\end{table*}

\begin{table*}[tp]
\centering
\small
\setlength{\tabcolsep}{4.2pt}
\begin{tabular}{llrrll}
\toprule
Model & Attribute & Act. AP & Prev. AP & $\Delta_{\mathrm{prev}}$ [95\% CI] & $\Delta_{\mathrm{text}}$ [95\% CI] \\
\midrule
OLMo & INTERNAL & .722 & .516 & .205 [.171, .241] & .063 [.028, .100] \\
 & SOCIAL & .679 & .370 & .309 [.259, .361] & .121 [.059, .174] \\
 & WORLD-STATE & .702 & .529 & .173 [.134, .215] & .086 [.041, .123] \\
\addlinespace[2pt]
Qwen & INTERNAL & .764 & .536 & .228 [.197, .259] & .036 [.010, .062] \\
 & SOCIAL & .400 & .186 & .214 [.164, .272] & .036 [-.015, .073] \\
 & WORLD-STATE & .765 & .567 & .198 [.167, .229] & .050 [.018, .079] \\
\addlinespace[2pt]
Gemma & INTERNAL & .814 & .594 & .220 [.187, .251] & .037 [.012, .061] \\
 & SOCIAL & .673 & .283 & .389 [.339, .441] & .108 [.042, .149] \\
 & WORLD-STATE & .777 & .540 & .236 [.209, .264] & .098 [.065, .125] \\
\addlinespace[2pt]
Mistral & INTERNAL & .779 & .564 & .215 [.180, .252] & .080 [.039, .116] \\
 & SOCIAL & .402 & .246 & .155 [.112, .210] & .061 [.010, .105] \\
 & WORLD-STATE & .786 & .587 & .200 [.168, .232] & .102 [.059, .139] \\
\bottomrule
\end{tabular}
\caption{Next-event attribute-specific scores and gains over prevalence and the larger primary text AP for each attribute. The text maximum is recomputed in each bootstrap replicate. All intervals are pointwise; these attribute-specific comparisons are exploratory and do not use the primary Macro-AP significance criterion.}
\label{tab:attribute-next-event}
\end{table*}

\subsection{Layer Selection and Horizon-Level Contrasts}

\begin{figure*}[tp]
\centering
\includegraphics[width=\textwidth]{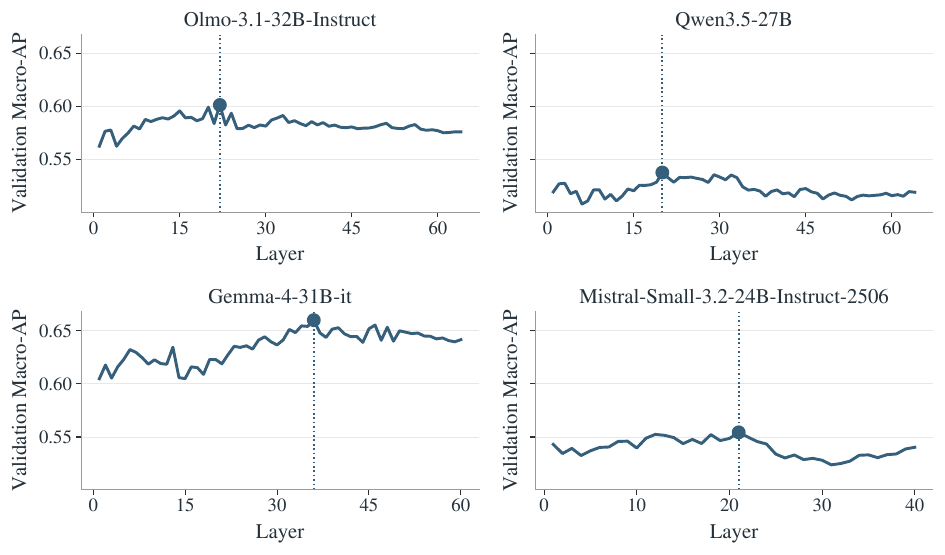}
\caption{Layer selection by mean event-end validation Macro-AP across eligible horizons. Dots and vertical lines identify the selected layers. The prompt-end anchor is excluded from this criterion, and test results do not alter these choices.}
\label{fig:validation-layers}
\end{figure*}

\begin{figure*}[tp]
\centering
\includegraphics[width=\textwidth]{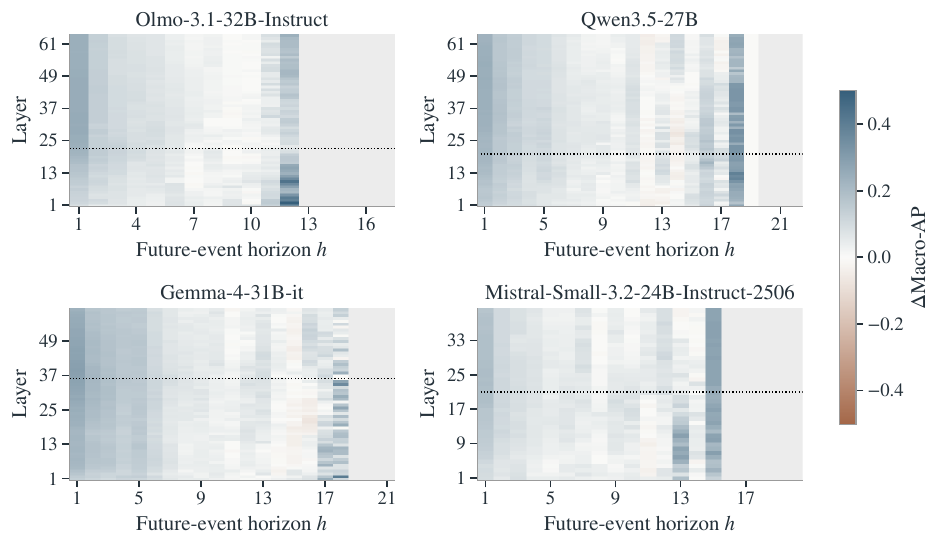}
\caption{Exploratory test Macro-AP gain over prevalence by layer and horizon. Dotted lines mark the validation-selected layers. Gray cells have no displayed estimate. No additional layer selection or significance testing uses this heatmap.}
\label{fig:all-layer-test}
\end{figure*}

Figure~\ref{fig:validation-layers} shows the layerwise validation criterion underlying the selections in Table~\ref{tab:probe-settings}. Figure~\ref{fig:all-layer-test} displays test gains across layers as an exploratory diagnostic, without changing those selections. Tables~\ref{tab:horizons-olmo}--\ref{tab:horizons-mistral} report the primary contrasts and corrected $p$-values; wide intervals at sparsely represented horizons should be read alongside their sample counts.

\begin{table*}[tp]
\centering
\small
\setlength{\tabcolsep}{4.2pt}
\begin{tabular}{rrlrlrr}
\toprule
$h$ & Stories/anchors & $\Delta_{\mathrm{prev}}$ [95\% CI] & $p_{\mathrm{prev}}$ & $\Delta_{\mathrm{text}}$ [95\% CI] & $p_{\mathrm{full}}$ & $p_{\mathrm{chunk}}$ \\
\midrule
1 & 100/825 & .229 [.203, .256] & .0011 & .090 [.061, .115] & .012 & .012 \\
2 & 100/725 & .141 [.114, .174] & .0011 & .027 [-.002, .053] & .952 & 1.000 \\
3 & 100/625 & .120 [.089, .158] & .0011 & .027 [-.000, .052] & 1.000 & 1.000 \\
4 & 100/525 & .094 [.064, .129] & .0011 & .026 [-.013, .052] & 1.000 & 1.000 \\
5 & 98/425 & .074 [.045, .112] & .0011 & .006 [-.031, .041] & .864 & 1.000 \\
6 & 93/327 & .052 [.017, .107] & .199 & .035 [-.022, .080] & 1.000 & 1.000 \\
7 & 78/234 & .021 [-.012, .075] & 1.000 & -.023 [-.085, .030] & 1.000 & 1.000 \\
8 & 66/156 & .040 [-.001, .111] & .823 & -.031 [-.108, .040] & 1.000 & 1.000 \\
9 & 43/90 & .005 [-.029, .093] & 1.000 & -.045 [-.141, .054] & 1.000 & 1.000 \\
10 & 24/47 & .023 [-.032, .145] & 1.000 & -.039 [-.164, .068] & 1.000 & 1.000 \\
11 & 12/23 & .045 [-.006, .198] & 1.000 & -.064 [-.243, .105] & 1.000 & 1.000 \\
\bottomrule
\end{tabular}
\caption{OLMo horizon-level Macro-AP contrasts. $\Delta_{\mathrm{prev}}$ is activation minus prevalence; $\Delta_{\mathrm{text}}$ is activation minus the larger of the two primary text scores, with the maximum recomputed in every bootstrap replicate. All CIs are pointwise. $p_{\mathrm{prev}}$ is Holm-adjusted within the model across eligible horizons; $p_{\mathrm{full}}$ and $p_{\mathrm{chunk}}$ use the shared 120-test primary text family. Text superiority requires both positive pairwise differences and both adjusted $p<.05$.}
\label{tab:horizons-olmo}
\end{table*}

\begin{table*}[tp]
\centering
\small
\setlength{\tabcolsep}{4.2pt}
\begin{tabular}{rrlrlrr}
\toprule
$h$ & Stories/anchors & $\Delta_{\mathrm{prev}}$ [95\% CI] & $p_{\mathrm{prev}}$ & $\Delta_{\mathrm{text}}$ [95\% CI] & $p_{\mathrm{full}}$ & $p_{\mathrm{chunk}}$ \\
\midrule
1 & 100/1254 & .213 [.189, .240] & .0018 & .041 [.020, .060] & .012 & .012 \\
2 & 100/1154 & .158 [.134, .187] & .0018 & .016 [-.009, .035] & 1.000 & 1.000 \\
3 & 100/1054 & .139 [.114, .171] & .0018 & .008 [-.019, .032] & 1.000 & 1.000 \\
4 & 100/954 & .102 [.077, .137] & .0018 & -.017 [-.043, .007] & 1.000 & 1.000 \\
5 & 100/854 & .114 [.083, .152] & .0018 & -.002 [-.028, .021] & 1.000 & 1.000 \\
6 & 98/754 & .075 [.051, .110] & .0018 & -.018 [-.045, .009] & 1.000 & 1.000 \\
7 & 97/656 & .063 [.041, .093] & .0018 & -.014 [-.041, .008] & 1.000 & 1.000 \\
8 & 95/559 & .045 [.015, .089] & .200 & -.034 [-.072, .008] & 1.000 & 1.000 \\
9 & 91/464 & .062 [.028, .109] & .044 & -.006 [-.047, .032] & 1.000 & 1.000 \\
10 & 83/373 & .024 [-.003, .068] & 1.000 & -.052 [-.094, -.012] & 1.000 & 1.000 \\
11 & 75/290 & .068 [.026, .134] & .200 & -.034 [-.093, .034] & 1.000 & 1.000 \\
12 & 68/215 & -.023 [-.052, .026] & 1.000 & -.053 [-.118, -.003] & 1.000 & 1.000 \\
13 & 48/147 & .064 [.007, .147] & .537 & -.031 [-.106, .035] & 1.000 & 1.000 \\
14 & 39/99 & -.016 [-.054, .060] & 1.000 & -.047 [-.116, .027] & 1.000 & 1.000 \\
15 & 28/60 & .043 [.005, .147] & 1.000 & .012 [-.082, .100] & 1.000 & 1.000 \\
16 & 16/32 & .138 [.046, .288] & .265 & .090 [-.085, .217] & 1.000 & 1.000 \\
17 & 10/16 & .104 [-.024, .302] & 1.000 & .093 [-.094, .239] & 1.000 & 1.000 \\
18 & 5/6 & .301 [.001, .556] & .200 & .199 [-.260, .396] & 1.000 & 1.000 \\
\bottomrule
\end{tabular}
\caption{Qwen horizon-level Macro-AP contrasts. $\Delta_{\mathrm{prev}}$ is activation minus prevalence; $\Delta_{\mathrm{text}}$ is activation minus the larger of the two primary text scores, with the maximum recomputed in every bootstrap replicate. All CIs are pointwise. $p_{\mathrm{prev}}$ is Holm-adjusted within the model across eligible horizons; $p_{\mathrm{full}}$ and $p_{\mathrm{chunk}}$ use the shared 120-test primary text family. Text superiority requires both positive pairwise differences and both adjusted $p<.05$.}
\label{tab:horizons-qwen}
\end{table*}

\begin{table*}[tp]
\centering
\small
\setlength{\tabcolsep}{4.2pt}
\begin{tabular}{rrlrlrr}
\toprule
$h$ & Stories/anchors & $\Delta_{\mathrm{prev}}$ [95\% CI] & $p_{\mathrm{prev}}$ & $\Delta_{\mathrm{text}}$ [95\% CI] & $p_{\mathrm{full}}$ & $p_{\mathrm{chunk}}$ \\
\midrule
1 & 100/1237 & .282 [.259, .305] & .0017 & .081 [.055, .102] & .012 & .012 \\
2 & 100/1137 & .215 [.191, .239] & .0017 & .012 [-.006, .028] & 1.000 & .327 \\
3 & 100/1037 & .181 [.156, .209] & .0017 & -.014 [-.035, .006] & 1.000 & 1.000 \\
4 & 100/937 & .160 [.134, .190] & .0017 & -.022 [-.044, -.001] & 1.000 & 1.000 \\
5 & 100/837 & .155 [.129, .188] & .0017 & -.020 [-.043, .005] & 1.000 & 1.000 \\
6 & 100/737 & .104 [.075, .137] & .0017 & -.043 [-.067, -.019] & 1.000 & .089 \\
7 & 100/637 & .087 [.055, .123] & .0017 & -.026 [-.056, -.001] & 1.000 & 1.000 \\
8 & 97/537 & .064 [.036, .099] & .0020 & -.020 [-.058, .009] & 1.000 & 1.000 \\
9 & 94/440 & .060 [.031, .099] & .0090 & -.015 [-.053, .020] & 1.000 & 1.000 \\
10 & 88/346 & .033 [.004, .076] & .553 & -.041 [-.090, .002] & 1.000 & 1.000 \\
11 & 77/258 & .020 [-.013, .071] & 1.000 & -.024 [-.073, .024] & 1.000 & 1.000 \\
12 & 67/181 & .064 [.025, .119] & .109 & .021 [-.049, .067] & 1.000 & 1.000 \\
13 & 47/114 & .086 [.033, .168] & .141 & .038 [-.045, .121] & 1.000 & 1.000 \\
14 & 29/67 & .037 [-.004, .108] & 1.000 & -.029 [-.106, .043] & 1.000 & 1.000 \\
15 & 19/38 & .010 [-.038, .142] & 1.000 & -.051 [-.141, .071] & 1.000 & 1.000 \\
16 & 9/19 & .019 [-.026, .190] & 1.000 & -.095 [-.236, .070] & 1.000 & 1.000 \\
17 & 5/10 & .077 [-.070, .422] & 1.000 & -.017 [-.359, .267] & 1.000 & 1.000 \\
\bottomrule
\end{tabular}
\caption{Gemma horizon-level Macro-AP contrasts. $\Delta_{\mathrm{prev}}$ is activation minus prevalence; $\Delta_{\mathrm{text}}$ is activation minus the larger of the two primary text scores, with the maximum recomputed in every bootstrap replicate. All CIs are pointwise. $p_{\mathrm{prev}}$ is Holm-adjusted within the model across eligible horizons; $p_{\mathrm{full}}$ and $p_{\mathrm{chunk}}$ use the shared 120-test primary text family. Text superiority requires both positive pairwise differences and both adjusted $p<.05$.}
\label{tab:horizons-gemma}
\end{table*}

\begin{table*}[tp]
\centering
\small
\setlength{\tabcolsep}{4.2pt}
\begin{tabular}{rrlrlrr}
\toprule
$h$ & Stories/anchors & $\Delta_{\mathrm{prev}}$ [95\% CI] & $p_{\mathrm{prev}}$ & $\Delta_{\mathrm{text}}$ [95\% CI] & $p_{\mathrm{full}}$ & $p_{\mathrm{chunk}}$ \\
\midrule
1 & 100/963 & .190 [.168, .216] & .0014 & .081 [.054, .105] & .012 & .012 \\
2 & 100/863 & .112 [.091, .138] & .0014 & .008 [-.020, .029] & 1.000 & 1.000 \\
3 & 100/763 & .091 [.069, .119] & .0014 & -.014 [-.050, .015] & 1.000 & 1.000 \\
4 & 100/663 & .075 [.048, .107] & .0014 & -.019 [-.048, .009] & 1.000 & 1.000 \\
5 & 98/563 & .044 [.018, .078] & .053 & -.057 [-.088, -.026] & .165 & .045 \\
6 & 98/465 & .065 [.037, .104] & .0040 & -.015 [-.048, .016] & 1.000 & 1.000 \\
7 & 95/367 & .048 [.010, .094] & .197 & -.015 [-.067, .032] & 1.000 & 1.000 \\
8 & 83/272 & .022 [-.010, .077] & .969 & -.026 [-.082, .032] & 1.000 & 1.000 \\
9 & 61/189 & .066 [.015, .140] & .284 & .031 [-.060, .112] & 1.000 & 1.000 \\
10 & 45/128 & .074 [.032, .146] & .175 & .017 [-.047, .091] & 1.000 & 1.000 \\
11 & 35/83 & .068 [.025, .151] & .284 & .006 [-.083, .094] & 1.000 & 1.000 \\
12 & 25/48 & .088 [.023, .187] & .284 & -.063 [-.158, .018] & 1.000 & 1.000 \\
13 & 10/23 & .053 [.011, .211] & .969 & .019 [-.118, .186] & 1.000 & 1.000 \\
14 & 7/13 & .006 [-.027, .236] & .969 & -.047 [-.273, .180] & 1.000 & 1.000 \\
\bottomrule
\end{tabular}
\caption{Mistral horizon-level Macro-AP contrasts. $\Delta_{\mathrm{prev}}$ is activation minus prevalence; $\Delta_{\mathrm{text}}$ is activation minus the larger of the two primary text scores, with the maximum recomputed in every bootstrap replicate. All CIs are pointwise. $p_{\mathrm{prev}}$ is Holm-adjusted within the model across eligible horizons; $p_{\mathrm{full}}$ and $p_{\mathrm{chunk}}$ use the shared 120-test primary text family. Text superiority requires both positive pairwise differences and both adjusted $p<.05$.}
\label{tab:horizons-mistral}
\end{table*}

\subsection{Time-Resolved Analyses and Inference Resolution}
\label{app:time-results}

\begin{figure*}[tp]
\centering
\includegraphics[width=\textwidth]{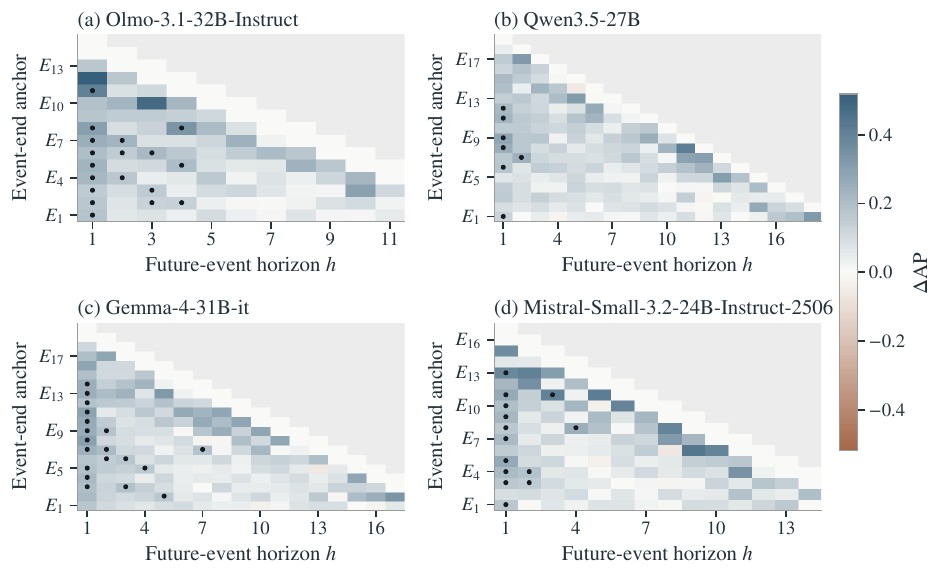}
\caption{Time-resolved activation Macro-AP gain over prevalence, indexed by event-end anchor and future-event horizon. Dots identify positive differences with within-model secondary-family Holm-adjusted $p<.05$. Gray cells have no displayed estimate. Missing significance markers do not establish absence of predictive information.}
\label{fig:time-activation}
\end{figure*}

\begin{figure*}[tp]
\centering
\includegraphics[width=\textwidth]{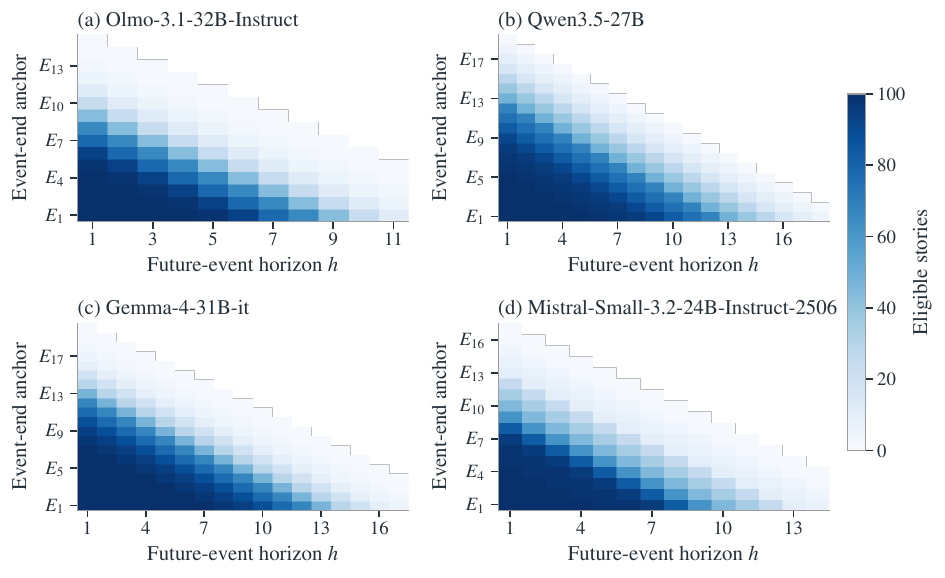}
\caption{Eligible test-story counts for the time-resolved analyses. Later anchors and longer horizons have fewer contributing stories; these counts should accompany interpretation of the effect heatmaps.}
\label{fig:time-support}
\end{figure*}

Figure~\ref{fig:time-activation} conditions on event-end anchor ordinal as well as future-event horizon. Positive activation-minus-prevalence differences survive the secondary within-model Holm correction in 18/110 cells for OLMo, 7/189 for Qwen, 19/204 for Gemma, and 14/147 for Mistral. The respective bootstrap counts are 11,000, 18,900, 20,400, and 14,700. These cell counts are not directly comparable measures of model capability: available cells and story counts differ. Figure~\ref{fig:time-support} shows the contributing story counts, which fall toward the edges of each grid. All reported test cells exclude the prompt-end anchor $E_0$.

\begin{figure*}[tp]
\centering
\includegraphics[width=\textwidth]{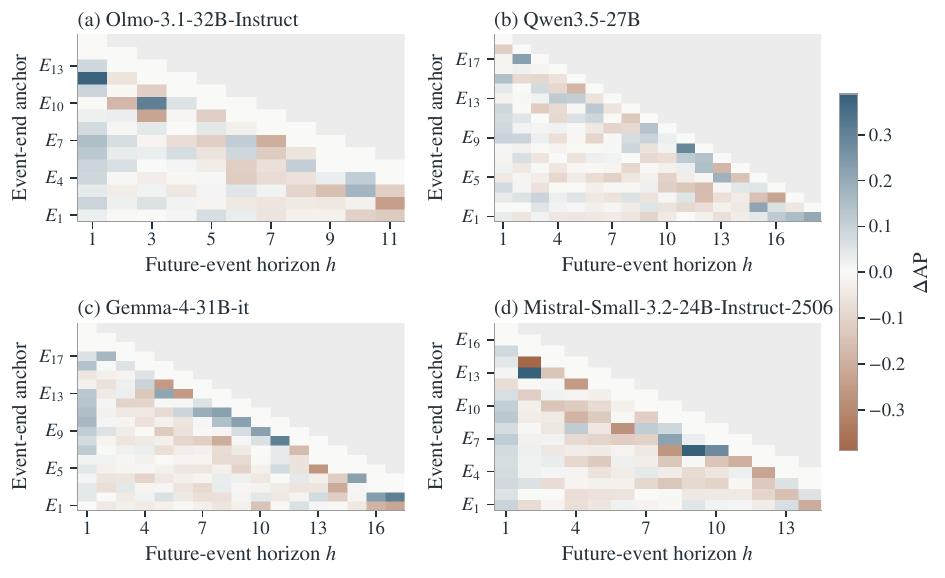}
\caption{Descriptive time-resolved activation Macro-AP minus the larger primary text Macro-AP. Gray cells have no displayed estimate. The recorded bootstrap resolution cannot reach the first Holm threshold of the 1,300-test secondary family (Appendix~\ref{app:time-results}); lack of significance is not evidence of no advantage.}
\label{fig:time-text}
\end{figure*}

The activation--text analysis in Figure~\ref{fig:time-text} requires an additional qualification. Its correction family contains 1,300 pairwise tests, but each test uses 10,000 bootstrap replicates. Under the recorded plus-one estimator, the smallest attainable raw $p$-value is $1/10{,}001\approx.000100$, exceeding the first Holm threshold $.05/1{,}300\approx.0000385$. Consequently, this implementation cannot reject any hypothesis in that secondary family, irrespective of the observed effect. We report the time-resolved text contrasts descriptively and do not interpret their lack of significance as evidence of no predictive advantage. This resolution constraint does not apply to the primary horizon-level text family, which has 120 tests and an initial threshold $.05/120\approx.000417$. No additional resampling or revised testing threshold is introduced in reporting these results.

\section{Additional Steering Results}
\label{app:steering-results}

\subsection{Directional Effects and the Random Control}

The OLMo experiment contains all 2,400 planned outputs: 100 matched seeds, four conditions, and six strengths. Tables~\ref{tab:steering-no} and~\ref{tab:steering-random} report SOCIAL contrasts against no steering and the random direction, respectively. Bidirectional support against no steering occurs only at $k=16$ and $32$. At $k=4$, both directional 95\% intervals exclude zero, but both Holm-adjusted $p$-values are .0549. The pointwise intervals therefore do not meet the study's corrected significance criterion. At $k=8$, only negative steering passes the criterion against no steering; both directions pass against random.

The random direction is not behaviorally neutral. Its SOCIAL differences from no steering are $-6.58$ percentage points at $k=16$ (95\% CI [$-11.16$, $-1.92$]; unadjusted $p=.0072$) and $-7.81$ at $k=32$ ([$-12.85$, $-2.66$]; $p=.0036$). Nevertheless, the probe-guided contrasts against random retain the intended signs at both strengths: at $k=16$, they are $+23.70$ and $-17.78$ points; at $k=32$, $+33.88$ and $-24.88$, all with supporting-family Holm-adjusted $p<.001$. A single orthogonal random direction offers a limited perturbation control, not a characterization of effects across arbitrary directions. Corrections remain within each strength, and the full sweep is descriptive.

\subsection{Generation Quality and Event Counts}

\begin{figure*}[tp]
\centering
\includegraphics[width=\textwidth]{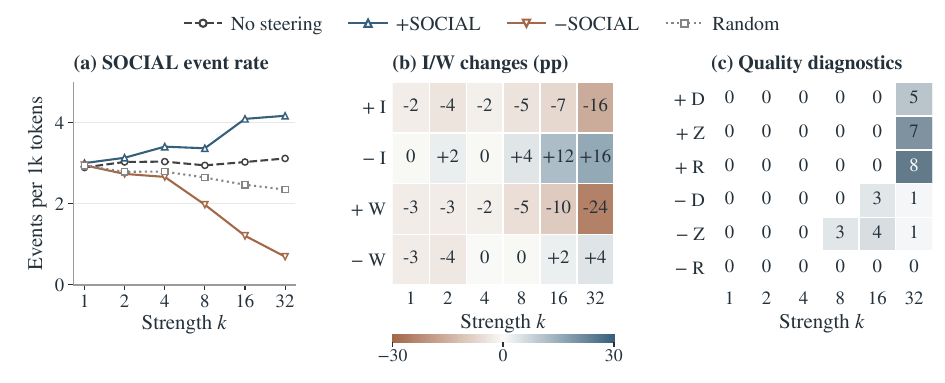}
\caption{Supplementary OLMo steering diagnostics ($100$ matched seeds per condition and strength). (a) Mean SOCIAL events per 1,000 tokens. (b) INTERNAL (I) and WORLD-STATE (W) differences against no steering, rounded to percentage points. (c) Degeneration (D), zero-event (Z), and disruptive-repetition (R) counts for positive/negative steering. No steering has zero flags; random has one R flag at $k=32$ and no D/Z flags. Flags can overlap. All outputs are retained. These panels are descriptive summaries without new significance tests.}
\label{fig:steering-diagnostics}
\end{figure*}

Figure~\ref{fig:steering-diagnostics} groups the supplementary event-count, attribute-change, and quality diagnostics. Table~\ref{tab:steering-quality} expands its positive/negative-steering count matrix (panel c) to every condition and strength, including both controls. No-steering outputs have no degeneration or zero-event flags across the sweep. At $k=16$, positive steering also has none; negative steering has 3/100 degenerate and 4/100 zero-event outputs. At $k=32$, positive steering has 8/100 disruptive-repetition flags, 5/100 degeneration flags, and 7/100 zero-event outputs. These indicators can overlap and are not summed into a single failure count. Negative steering at $k=32$ has one degeneration flag and one zero-event output. No output ends by EOS before 32 content tokens; three positive-steering outputs at $k=32$ reach the length limit, and no other condition--strength cell has a length-limit hit. Automatic annotation flags are diagnostics rather than an independent human evaluation of story quality.

The primary score includes every output and assigns zero to zero-event stories. Table~\ref{tab:steering-sensitivity} supplies complementary summaries at $k=16$ and $32$. At $k=16$, mean SOCIAL event counts are 3.61 without steering, 4.61 with positive steering, and 1.44 with negative steering. At $k=32$, they are 3.67, 4.83, and .77. SOCIAL counts per 1,000 tokens change in the same directions (Figure~\ref{fig:steering-diagnostics}a), even though positive steering at $k=32$ also lowers mean total events from 9.12 to 6.64. Thus, the proportion changes accompany changes in label counts and are not explained solely by the denominator. These count summaries have no additional inferential tests.

Restricting descriptively to seeds with events in all four conditions leaves 96 seeds at $k=16$ and 92 at $k=32$. The SOCIAL ratios retain the intended ordering (positive $>$ no steering $>$ negative). This subset is defined by an outcome affected by the intervention; it is not a replacement for the all-output analysis or an independent causal robustness test.

\begin{table*}[tp]
\centering
\small
\setlength{\tabcolsep}{4.2pt}
\begin{tabular}{r l r l r}
\toprule
$k$ & $+\mathrm{SOCIAL}$: $\Delta$ [95\% CI] & $p_{+}$ & $-\mathrm{SOCIAL}$: $\Delta$ [95\% CI] & $p_{-}$ \\
\midrule
1 & +2.36 [-2.09, 7.00] & .645 & +1.13 [-3.02, 5.18] & .645 \\
2 & +1.58 [-2.93, 6.06] & .498 & -2.36 [-6.15, 1.37] & .465 \\
4 & +5.85 [.84, 10.95] & .055 & -5.62 [-10.47, -.68] & .055 \\
8 & +3.60 [-1.28, 8.52] & .153 & -13.69 [-18.53, -8.74] & $<.001$ \\
16 & +17.13 [12.45, 21.88] & $<.001$ & -24.36 [-28.89, -19.84] & $<.001$ \\
32 & +26.07 [18.76, 33.41] & $<.001$ & -32.70 [-37.18, -28.03] & $<.001$ \\
\bottomrule
\end{tabular}
\caption{OLMo SOCIAL contrasts against no steering ($n=100$ matched seeds at each strength). Differences and pointwise bootstrap intervals are in percentage points. Two-sided sign-flip $p$-values are Holm-adjusted across the two directional comparisons within each strength; no correction is applied across strengths.}
\label{tab:steering-no}
\end{table*}

\begin{table*}[tp]
\centering
\small
\setlength{\tabcolsep}{4.2pt}
\begin{tabular}{r l r l r}
\toprule
$k$ & $+\mathrm{SOCIAL}$: $\Delta$ [95\% CI] & $p_{+}$ & $-\mathrm{SOCIAL}$: $\Delta$ [95\% CI] & $p_{-}$ \\
\midrule
1 & +0.81 [-3.81, 5.44] & 1.000 & -0.43 [-5.02, 4.24] & 1.000 \\
2 & +3.53 [-.58, 7.62] & .205 & -0.42 [-4.56, 3.73] & .845 \\
4 & +8.60 [3.77, 13.45] & .0016 & -2.86 [-7.81, 2.14] & .266 \\
8 & +6.65 [1.91, 11.14] & .0063 & -10.64 [-15.14, -6.05] & $<.001$ \\
16 & +23.70 [18.73, 28.56] & $<.001$ & -17.78 [-21.85, -13.63] & $<.001$ \\
32 & +33.88 [27.43, 40.17] & $<.001$ & -24.88 [-29.01, -20.68] & $<.001$ \\
\bottomrule
\end{tabular}
\caption{OLMo SOCIAL contrasts against the fixed orthogonal random direction ($n=100$ matched seeds at each strength). Differences and pointwise bootstrap intervals are in percentage points. Two-sided sign-flip $p$-values are Holm-adjusted across the two directional comparisons within each strength; no correction is applied across strengths.}
\label{tab:steering-random}
\end{table*}

\begin{table*}[tp]
\centering
\small
\setlength{\tabcolsep}{4.2pt}
\begin{tabular}{rlrrrrrrr}
\toprule
$k$ & Condition & D & Z & R & Limit & Tokens & Events & Rep-4 (\%) \\
\midrule
1 & No steering & 0 & 0 & 0 & 0 & 1218.1 & 9.06 & 0.92 \\
 & $+\mathrm{SOCIAL}$ & 0 & 0 & 0 & 0 & 1224.7 & 8.98 & 1.05 \\
 & $-\mathrm{SOCIAL}$ & 0 & 0 & 0 & 0 & 1197.5 & 8.91 & 0.94 \\
 & Random & 0 & 0 & 0 & 0 & 1157.8 & 8.40 & 0.82 \\
\addlinespace[2pt]
2 & No steering & 0 & 0 & 0 & 0 & 1205.8 & 9.11 & 0.91 \\
 & $+\mathrm{SOCIAL}$ & 0 & 0 & 0 & 0 & 1149.1 & 8.61 & 0.91 \\
 & $-\mathrm{SOCIAL}$ & 0 & 0 & 0 & 0 & 1222.8 & 8.80 & 0.90 \\
 & Random & 0 & 0 & 0 & 0 & 1206.2 & 8.79 & 0.99 \\
\addlinespace[2pt]
4 & No steering & 0 & 0 & 0 & 0 & 1201.8 & 8.98 & 0.92 \\
 & $+\mathrm{SOCIAL}$ & 0 & 0 & 0 & 0 & 1196.0 & 8.80 & 0.91 \\
 & $-\mathrm{SOCIAL}$ & 0 & 0 & 0 & 0 & 1173.8 & 8.77 & 0.88 \\
 & Random & 0 & 0 & 0 & 0 & 1212.0 & 8.87 & 0.94 \\
\addlinespace[2pt]
8 & No steering & 0 & 0 & 0 & 0 & 1192.2 & 8.82 & 0.90 \\
 & $+\mathrm{SOCIAL}$ & 0 & 0 & 0 & 0 & 1185.6 & 8.96 & 0.99 \\
 & $-\mathrm{SOCIAL}$ & 0 & 3 & 0 & 0 & 1229.2 & 9.15 & 0.89 \\
 & Random & 0 & 0 & 0 & 0 & 1229.2 & 8.91 & 0.94 \\
\addlinespace[2pt]
16 & No steering & 0 & 0 & 0 & 0 & 1211.7 & 9.14 & 0.96 \\
 & $+\mathrm{SOCIAL}$ & 0 & 0 & 0 & 0 & 1149.6 & 8.19 & 1.21 \\
 & $-\mathrm{SOCIAL}$ & 3 & 4 & 0 & 0 & 1183.8 & 9.11 & 0.89 \\
 & Random & 0 & 0 & 0 & 0 & 1201.4 & 9.00 & 0.98 \\
\addlinespace[2pt]
32 & No steering & 0 & 0 & 0 & 0 & 1205.9 & 9.12 & 0.94 \\
 & $+\mathrm{SOCIAL}$ & 5 & 7 & 8 & 3 & 1172.8 & 6.64 & 3.57 \\
 & $-\mathrm{SOCIAL}$ & 1 & 1 & 0 & 0 & 1134.2 & 9.08 & 1.55 \\
 & Random & 0 & 0 & 1 & 0 & 1287.8 & 8.91 & 1.31 \\
\bottomrule
\end{tabular}
\caption{OLMo generation diagnostics ($100$ outputs per row). D, Z, and R count annotator-flagged degeneration, zero-event outputs, and annotator-flagged disruptive repetition. Limit counts outputs reaching the 4,000-token limit. Tokens and Events are per-story means. Rep-4 is the mean repeated-token-4-gram fraction, in percent. Flags may overlap. No output ends by EOS before 32 content tokens. All outputs enter the primary score.}
\label{tab:steering-quality}
\end{table*}

\begin{table*}[tp]
\centering
\small
\setlength{\tabcolsep}{4.2pt}
\begin{tabular}{rlrrrrrr}
\toprule
$k$ & Condition & Score (\%) & Common (\%) & $N_{\mathrm{common}}$ & S count & S/1k tokens & Events \\
\midrule
16 & No steering & 39.46 & 39.16 & 96 & 3.61 & 3.03 & 9.14 \\
 & $+\mathrm{SOCIAL}$ & 56.59 & 56.71 & 96 & 4.61 & 4.10 & 8.19 \\
 & $-\mathrm{SOCIAL}$ & 15.10 & 15.73 & 96 & 1.44 & 1.21 & 9.11 \\
 & Random & 32.89 & 33.02 & 96 & 2.92 & 2.47 & 9.00 \\
\addlinespace[2pt]
32 & No steering & 41.03 & 40.51 & 92 & 3.67 & 3.12 & 9.12 \\
 & $+\mathrm{SOCIAL}$ & 67.10 & 72.25 & 92 & 4.83 & 4.17 & 6.64 \\
 & $-\mathrm{SOCIAL}$ & 8.34 & 8.63 & 92 & 0.77 & 0.70 & 9.08 \\
 & Random & 33.22 & 33.23 & 92 & 2.99 & 2.35 & 8.91 \\
\bottomrule
\end{tabular}
\caption{Descriptive SOCIAL sensitivity summaries at the two strengths with bidirectional support against no steering. Score includes all 100 outputs with zero-event scores set to zero. Common is the mean ratio restricted to seeds whose outputs have events under all four conditions. S count, S/1k tokens, and Events are per-story means over all outputs. The common-eventful subset is post-intervention; it does not define an additional causal test.}
\label{tab:steering-sensitivity}
\end{table*}

\subsection{Supplementary Attributes and Scope of the Intervention}

\begin{figure*}[tp]
\centering
\includegraphics[width=\textwidth]{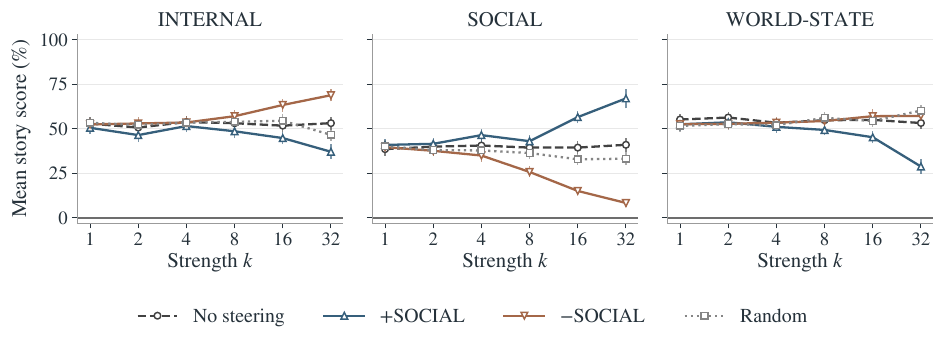}
\caption{Absolute mean per-story scores for OLMo under all four steering conditions ($n=100$ per condition and strength). Whiskers are pointwise 95\% seed-bootstrap intervals. Scores include zero-event outputs with value zero; multi-label attribute scores need not sum to one. Curves summarize the strength sweep descriptively and do not imply monotonic effects.}
\label{fig:steering-attributes}
\end{figure*}

Figure~\ref{fig:steering-diagnostics}b summarizes the supplementary INTERNAL and WORLD-STATE effects across the entire sweep; its integer cell labels are rounded for legibility and carry no significance markers. Figure~\ref{fig:steering-attributes} shows absolute scores for all three attributes. Table~\ref{tab:steering-other-attributes} reports INTERNAL and WORLD-STATE contrasts against no steering across all strengths, using their exploratory within-strength correction family. At $k=16$, positive SOCIAL steering reduces INTERNAL by 6.97 percentage points and WORLD-STATE by 9.76, while negative steering raises INTERNAL by 11.62. At $k=32$, the respective positive-steering changes are $-16.09$ and $-24.46$, and negative steering raises INTERNAL by 15.70. These contrasts have exploratory Holm-adjusted $p<.05$; the WORLD-STATE increases under negative steering at these strengths do not. Because attributes are multi-label, these shifts are not forced by a sum-to-one constraint.

The effects therefore extend beyond the targeted SOCIAL attribute, indicating a broader change in event composition. These sustained interventions do not establish selective control over individual future events or a naturally operative planning mechanism.

\begin{table*}[tp]
\centering
\small
\setlength{\tabcolsep}{4.2pt}
\begin{tabular}{rllrlr}
\toprule
$k$ & Outcome & $+\mathrm{SOCIAL}$: $\Delta$ [95\% CI] & $p_{+}$ & $-\mathrm{SOCIAL}$: $\Delta$ [95\% CI] & $p_{-}$ \\
\midrule
1 & INTERNAL & -2.34 [-6.66, 1.97] & .881 & -0.46 [-4.93, 4.12] & .881 \\
 & WORLD-STATE & -2.71 [-7.16, 1.65] & .881 & -2.64 [-6.82, 1.49] & .881 \\
\addlinespace[2pt]
2 & INTERNAL & -4.23 [-8.86, .22] & .277 & +2.30 [-2.43, 6.88] & .431 \\
 & WORLD-STATE & -2.68 [-6.88, 1.49] & .431 & -3.56 [-7.82, .88] & .336 \\
\addlinespace[2pt]
4 & INTERNAL & -2.26 [-6.87, 2.51] & 1.000 & -0.26 [-4.61, 4.11] & 1.000 \\
 & WORLD-STATE & -2.31 [-6.54, 1.93] & 1.000 & +0.03 [-4.08, 4.08] & 1.000 \\
\addlinespace[2pt]
8 & INTERNAL & -4.53 [-10.29, 1.28] & .389 & +3.94 [-1.36, 9.40] & .389 \\
 & WORLD-STATE & -5.46 [-9.43, -1.44] & .035 & -0.44 [-4.50, 3.61] & .835 \\
\addlinespace[2pt]
16 & INTERNAL & -6.97 [-12.30, -1.63] & .026 & +11.62 [6.98, 16.22] & $<.001$ \\
 & WORLD-STATE & -9.76 [-13.95, -5.55] & $<.001$ & +2.08 [-3.24, 7.22] & .439 \\
\addlinespace[2pt]
32 & INTERNAL & -16.09 [-21.50, -10.61] & $<.001$ & +15.70 [10.96, 20.45] & $<.001$ \\
 & WORLD-STATE & -24.46 [-29.25, -19.78] & $<.001$ & +4.01 [-.70, 8.53] & .093 \\
\bottomrule
\end{tabular}
\caption{Exploratory INTERNAL and WORLD-STATE contrasts against no steering, in percentage points, retaining all 100 outputs per condition. CIs are pointwise. Two-sided sign-flip $p$-values use one four-comparison Holm family within each strength (two attributes by two directions), without correction across strengths.}
\label{tab:steering-other-attributes}
\end{table*}

\clearpage

\section{Event Annotation Prompt}
\label{app:annotation-prompt}

\subsection{Event Annotation Rubric}
\label{app:annotation-prompt}
\noindent The following reproduces the annotation instructions used in the experiment. Section headings and separator lines are adjusted for readability; the instructional wording is unchanged.
\par\medskip

\begin{tcblisting}{
  enhanced, breakable,
  listing only, listing engine=listings,
  title={Annotation instructions},
  fonttitle=\bfseries\upshape\footnotesize,
  colback=white, colbacktitle=black!12,
  coltitle=black, colframe=black!25,
  boxrule=0.45pt, arc=1.5pt,
  left=5pt, right=5pt, top=5pt, bottom=5pt,
  toptitle=3pt, bottomtitle=3pt,
  before skip=6pt, after skip=7pt,
  listing options={basicstyle=\ttfamily\small,breaklines=true,breakatwhitespace=true,columns=fullflexible,keepspaces=true}
}
You are annotating the major narrative events of a fictional story.

The purpose of this annotation is to represent the major narrative state transitions of the story at an abstract level.

IMPORTANT:

First identify the main narrative events WITHOUT using the taxonomy to decide what counts as an event.

Only after identifying the main events should you assign the abstract state-change attributes.

Do NOT segment the story sentence by sentence.
Do NOT extract every action.
Do NOT extract every possible state change.
\end{tcblisting}

\begin{tcblisting}{
  enhanced, breakable,
  listing only, listing engine=listings,
  title={A. Main-event extraction},
  fonttitle=\bfseries\upshape\footnotesize,
  colback=white, colbacktitle=black!12,
  coltitle=black, colframe=black!25,
  boxrule=0.45pt, arc=1.5pt,
  left=5pt, right=5pt, top=5pt, bottom=5pt,
  toptitle=3pt, bottomtitle=3pt,
  before skip=6pt, after skip=7pt,
  listing options={basicstyle=\ttfamily\small,breaklines=true,breakatwhitespace=true,columns=fullflexible,keepspaces=true}
}
RULE 1 - Identify major state transitions, not every state change.

Extract a passage as a main event when it establishes a meaningful new narrative state relative to what came before and constitutes a significant development in the progression of the story.

Small intermediate changes should not be extracted independently.


RULE 2 - Exclude local actions, descriptions, and intermediate steps.

Do not independently extract:

- minor physical actions
- incidental movements
- background description
- temporary emotional expressions
- intermediate actions that merely contribute to a larger event

If several small actions together constitute one larger narrative development, treat them as one event.


RULE 3 - Group tightly connected transitions into one main event.

If several changes directly constitute the same coherent narrative development and are closely connected temporally or causally, group them into ONE main event.

However, do NOT merge events merely because they occur next to each other.

If two developments independently establish distinct narrative states, extract them as separate events.

The number of state-change attributes does NOT determine the number of events.

One coherent narrative event may change multiple domains.


RULE 4 - Use complete sentence units.

Select the smallest number of COMPLETE consecutive sentences necessary to capture the main event.

Do NOT select a partial sentence, even when only part of the sentence expresses the relevant state transition.

If one complete sentence is sufficient, select that full sentence.

Use multiple complete consecutive sentences only when they are necessary to represent the same coherent narrative development.

Avoid including unnecessary neighboring sentences.


RULE 5 - Do not automatically include peripheral causes or consequences.

Do not include preceding causes or later consequences unless they are necessary components of the same main event.
\end{tcblisting}

\begin{tcblisting}{
  enhanced, breakable,
  listing only, listing engine=listings,
  title={B. Attempted transitions},
  fonttitle=\bfseries\upshape\footnotesize,
  colback=white, colbacktitle=black!12,
  coltitle=black, colframe=black!25,
  boxrule=0.45pt, arc=1.5pt,
  left=5pt, right=5pt, top=5pt, bottom=5pt,
  toptitle=3pt, bottomtitle=3pt,
  before skip=6pt, after skip=7pt,
  listing options={basicstyle=\ttfamily\small,breaklines=true,breakatwhitespace=true,columns=fullflexible,keepspaces=true}
}
An explicit attempt may be extracted as a main event when the attempted transition itself constitutes a major narrative development, even if the target state is not achieved.

Attempted / Realized is NOT part of the primary event taxonomy.

Record it only as metadata:

Attempted
or
Realized
\end{tcblisting}

\begin{tcblisting}{
  enhanced, breakable,
  listing only, listing engine=listings,
  title={C. Abstract state-change attributes},
  fonttitle=\bfseries\upshape\footnotesize,
  colback=white, colbacktitle=black!12,
  coltitle=black, colframe=black!25,
  boxrule=0.45pt, arc=1.5pt,
  left=5pt, right=5pt, top=5pt, bottom=5pt,
  toptitle=3pt, bottomtitle=3pt,
  before skip=6pt, after skip=7pt,
  listing options={basicstyle=\ttfamily\small,breaklines=true,breakatwhitespace=true,columns=fullflexible,keepspaces=true}
}
After identifying each main event, determine where the relevant narrative state change is instantiated.

An event may receive MORE THAN ONE attribute.


INTERNAL (I)

A meaningful state change within a character.

Examples:

- knowledge or belief change
- realization
- meaningful discovery
- decision or commitment
- meaningful emotional change
- change in self-understanding
- internal transformation


SOCIAL (S)

A meaningful change instantiated in interpersonal or social structure.

Examples:

- relationship change
- alliance or conflict
- role change
- status change
- social recognition
- authority
- participation in a social group or institution


WORLD-STATE (W)

A meaningful change instantiated in the external storyworld.

Examples:

- possession
- location
- bodily state
- object state
- environmental state
- physical condition
- technological or magical system state
- larger storyworld state
\end{tcblisting}

\begin{tcblisting}{
  enhanced, breakable,
  listing only, listing engine=listings,
  title={D. Multiple attributes},
  fonttitle=\bfseries\upshape\footnotesize,
  colback=white, colbacktitle=black!12,
  coltitle=black, colframe=black!25,
  boxrule=0.45pt, arc=1.5pt,
  left=5pt, right=5pt, top=5pt, bottom=5pt,
  toptitle=3pt, bottomtitle=3pt,
  before skip=6pt, after skip=7pt,
  listing options={
      basicstyle=\ttfamily\small,
      breaklines=true,
      breakatwhitespace=true,
      columns=fullflexible,
      keepspaces=true
    }
}
Multiple attributes do NOT imply multiple events.

For example:

"Marren discovered the hidden key and took it with her."

If this is one coherent narrative development, annotate it as ONE event:

Internal = 1
Social = 0
World-state = 1

The resulting sequence label is I+W.

Do NOT split the event merely because multiple attributes apply.
\end{tcblisting}

\begin{tcblisting}{
  enhanced, breakable,
  listing only, listing engine=listings,
  title={E. Important distinctions},
  fonttitle=\bfseries\upshape\footnotesize,
  colback=white, colbacktitle=black!12,
  coltitle=black, colframe=black!25,
  boxrule=0.45pt, arc=1.5pt,
  left=5pt, right=5pt, top=5pt, bottom=5pt,
  toptitle=3pt, bottomtitle=3pt,
  before skip=6pt, after skip=7pt,
  listing options={basicstyle=\ttfamily\small,breaklines=true,breakatwhitespace=true,columns=fullflexible,keepspaces=true}
}
Do not mechanically assign Internal merely because a perception verb or a word such as "discover" appears.

First determine whether the passage constitutes a main narrative event.
Then determine whether a meaningful epistemic or internal state change is part of that event.

Likewise, ordinary physical movement does not automatically constitute a World-state main event.

The movement must establish a meaningful narrative state transition.
\end{tcblisting}

\begin{tcblisting}{
  enhanced, breakable,
  listing only, listing engine=listings,
  title={F. Span requirement},
  fonttitle=\bfseries\upshape\footnotesize,
  colback=white, colbacktitle=black!12,
  coltitle=black, colframe=black!25,
  boxrule=0.45pt, arc=1.5pt,
  left=5pt, right=5pt, top=5pt, bottom=5pt,
  toptitle=3pt, bottomtitle=3pt,
  before skip=6pt, after skip=7pt,
  listing options={basicstyle=\ttfamily\small,breaklines=true,breakatwhitespace=true,columns=fullflexible,keepspaces=true}
}
For every event, copy the event span VERBATIM from the supplied story.

Do not paraphrase it.

Do not correct its grammar or punctuation.

The span must be a contiguous passage from the original story.

Preserve narrative order.
\end{tcblisting}

\begin{tcblisting}{
  enhanced, breakable,
  listing only, listing engine=listings,
  title={Output format},
  fonttitle=\bfseries\upshape\footnotesize,
  colback=white, colbacktitle=black!12,
  coltitle=black, colframe=black!25,
  boxrule=0.45pt, arc=1.5pt,
  left=5pt, right=5pt, top=5pt, bottom=5pt,
  toptitle=3pt, bottomtitle=3pt,
  before skip=6pt, after skip=7pt,
  listing options={basicstyle=\ttfamily\small,breaklines=true,breakatwhitespace=true,columns=fullflexible,keepspaces=true}
}
Return ONLY valid JSON.
Do not include markdown fences or explanatory text.
Use exactly this structure:

{
  "events": [
    {
      "event_number": 1,
      "span": "exact verbatim span",
      "internal": 0,
      "social": 0,
      "world_state": 1,
      "transition_status": "Realized"
    }
  ]
}

internal, social, and world_state must each be 0 or 1.

At least one of the three state-change attributes must be 1.

transition_status must be either:
"Attempted"
or
"Realized"

Attempted / Realized is metadata only and does NOT alter the I/S/W sequence label.
\end{tcblisting}

\subsection{Steering Annotation Extension}
\label{app:steering-prompt-extension}

For OLMo steering, the original event-selection and labeling
instructions are retained. The following instructions are appended
to the original rubric:

\begin{tcblisting}{
  enhanced, breakable,
  listing only, listing engine=listings,
  title={Additional steering annotation instructions},
  fonttitle=\bfseries\upshape\footnotesize,
  colback=white, colbacktitle=black!12,
  coltitle=black, colframe=black!25,
  boxrule=0.45pt, arc=1.5pt,
  left=5pt, right=5pt, top=5pt, bottom=5pt,
  toptitle=3pt, bottomtitle=3pt,
  listing options={
    basicstyle=\ttfamily\small,
    breaklines=true,
    breakatwhitespace=true,
    columns=fullflexible,
    keepspaces=true
  }
}
If no passage meets the main-event criteria above, return "events": [].
repetition: obvious disruptive repetition of phrases, sentences, or narrative loops in the story.
narrative_degeneration: unreadable/incoherent narrative, gibberish, or sustained off-task output.
A short but coherent ending or the absence of eligible events alone is not narrative degeneration.
The two quality flags do not change event selection or labels.
\end{tcblisting}

The JSON output example in the original rubric is extended with
two required top-level Boolean fields, \texttt{repetition} and
\texttt{narrative\_degeneration}. Event fields and their permitted
values remain unchanged. Both quality flags are returned even
when \texttt{events} is empty. No additional explanation or
\texttt{notes} field is requested.

\begin{tcblisting}{
  enhanced, breakable,
  listing only, listing engine=listings,
  title={Extended JSON output example},
  fonttitle=\bfseries\upshape\footnotesize,
  colback=white, colbacktitle=black!12,
  coltitle=black, colframe=black!25,
  boxrule=0.45pt, arc=1.5pt,
  left=5pt, right=5pt, top=5pt, bottom=5pt,
  toptitle=3pt, bottomtitle=3pt,
  listing options={
    basicstyle=\ttfamily\small,
    breaklines=true,
    breakatwhitespace=true,
    columns=fullflexible,
    keepspaces=true
  }
}
{
  "events": [
    {
      "event_number": 1,
      "span": "exact verbatim span",
      "internal": 0,
      "social": 0,
      "world_state": 1,
      "transition_status": "Realized"
    }
  ],
  "repetition": false,
  "narrative_degeneration": false
}
\end{tcblisting}

\end{document}